\documentclass{article}

\newcommand{\commentOut}[1]{} 

\newcommand{\mat}[1]{\mathbf{#1}}
\renewcommand{\vec}[1]{ \mathbf{#1} } 

\newcommand{\I}{\mat{I}}

\newcommand{\X}{\mat{X}}

\renewcommand{\S}{\mat{S}}
\newcommand{\K}{\mat{K}}

\newcommand{\f}{\vec{f}}
\newcommand{\g}{\vec{g}}

\newcommand{\m}{\vec{m}}

\newcommand{\x}{\vec{x}}
\newcommand{\y}{\vec{y}}
\newcommand{\z}{\vec{z}}

\usepackage{amsmath}

\newcommand{\Normal}{\mathcal{N}}

\numberwithin{equation}{section}

\newcommand{\map}{\text{\sc{map}}}

\allowdisplaybreaks[2]

\newcommand{\Ex}{\mathbb E}

\usepackage{xspace}

\let\epsilon\varepsilon

\usepackage{hyperref}
\usepackage[ruled,vlined]{algorithm2e}

\usepackage{tikz}
\usetikzlibrary{automata, positioning}
\usepackage{blkarray}

\usepackage[preprint]{neurips_2020}

\renewcommand{\vec}[1]{ \mathbf{#1} } 

\SetCommentSty{mycommfont}

\usepackage[utf8]{inputenc} 
\usepackage[T1]{fontenc}    
\usepackage{hyperref}       
\usepackage{url}            
\usepackage{booktabs}       
\usepackage{amsfonts}       
\usepackage{nicefrac}       
\usepackage{microtype}  
\usepackage{natbib} 

\usepackage{algorithmic}

\title{Approximate Bayesian Optimisation for Neural Networks}

\author{%
  Nadhir Hassen, \\
  MILA, Quebec, Canada\\
  \texttt{nadhir.hassen@mila.quebec} \\
   \And
   Irina Rish \\
   MILA, Quebec, Canada \\
   \texttt{irina.rish@mila.quebec}
}

\begin{document}

\maketitle

\begin{abstract}
 A body of work has been done to automate machine learning algorithms and to highlight the importance of model choice. Automating the process of choosing the best forecasting model and its corresponding parameters can result to improve a wide range of real-world applications. Bayesian optimisation (BO) uses a black-box optimisation methods to propose solutions according to an exploration-exploitation trade-off criterion through acquisition functions. BO framework imposes two key ingredients: a probabilistic surrogate model that consists of prior belief of the unknown dynamic of the model and an objective function that describes how optimal the model-fit. Choosing the best model and its associated hyperparameters can be very expensive, and is typically fit using Gaussian processes (GPs). However, since GPs scale cubically with the number of observations, it has been challenging to handle objectives whose optimisation requires many evaluations. In addition, most real-datasets are non-stationary which makes idealistic assumptions on surrogate models. The necessity to solve the analytical tractability and the computational feasibility in a stochastic fashion enables to ensure the efficiency and the applicability of Bayesian optimisation. In this paper we explore the use of approximate inference with Bayesian Neural Networks as an alternative to GPs to model distributions over functions. Our contribution is to provide a link between density-ratio estimation and class probability estimation based on approximate inference, this reformulation provides algorithm efficiency and tractability.

 \end{abstract}
 \newpage

\section{Introduction}
Bayesian Optimisation is a powerful methodology for optimising black-box functions that are generally noisy, expensive to evaluate or have not accessible derivatives. The BO framework consists to propose candidate solutions according to an \emph{acquisition function} that encodes an exploration-exploitation trade-off to optimise a cost function for various hyperparameter configurations. These parameter can correspond to learning rates, layer depths, width of the neural network or dropout rates during the training phase. We can formulate the problem of finding a global optimiser of an unknown objective function $\f$ as follow
\begin{align*}
    \x_{\star} = \arg \max_{\x\in \mathcal{X}} \f(\x),
\end{align*}
where $\mathcal{X}\in \mathbb{R}^d$ is the space of interest and $\x$ denotes the variable that corresponds to the hyperparameter setting which can be continuous, discrete or mixed. BO can be applied to larger space that involves categorical input or combinatorial search spaces with multiple categorical inputs. We assume that the black-box function $\f$ has no explicit form and can be evaluated at any arbitrary point $\x$ in the domain. Typically, the BO is based on two key components, the surrogate model that approximates $\f(\x)$ and the acquisition function that measures the utility of new input points by trading off exploration and exploitation. The surrogate model is a probabilistic model that identifies which acquisition function should be computed. The most common probabilistic model is the Gaussian process (GP) due to its flexibility and its ability to handle uncertainty estimates. The GP is characterised by its mean which captures the estimated value of the function and whose variance determines the level of uncertainty of the prediction for a specific location. In particular, the Bayes rule is applied to prescribe a prior belief over the possible objective functions and sequentially updates the posterior belief as the data are observed to query for the most promising objective based on the acquisition function $\f$. Surrogate BO models present some issues due to GPs idealistic assumptions. In fact, GPs present notably limitations due to the computational complexity which scale to $\mathcal{O}(N^3)$. They are poorly calibrated to real-datasets, that is, the stationary assumption assumes the GP kernel translation-invariant where noise processes are complex and heteroscedastic. In addition, they are not equipped to deal with conditional dependency structures \cite{gp_structures_jenatton17a}. Alternatively, we can express an acquisition function in a special form that does not impose analytical tractability constraint on the surrogate model. Expected Improvement (EI) presented by \cite{NIPS2011_86e8f7ab} plays the role of criterion that corresponds to utility-function. The improved target value which is unknown converges reasonably in practice. \citet{Mockus1978} demonstrate that EI function can be expressed as \emph{relative ratio} between two densities termed \emph{the tree-structured Parzen estimator} (TPE) \cite{bengio_bergstra_algoHPO}. This method which is based on density estimation can naturally deal with tree-structured inputs, discrete inputs, and scales linearly with the number of observations.

However these methods are not adaptive to non-stationary surrogate functions which may lead to numerical instability and scale poorly to higher dimensions. \cite{linking_cpe_pmlr-v48-menon16} and \cite{tiao2021-bore} exploit the link between density ratio estimation (DRE) and the EI function by class-probability estimation (CPE). This method has the advantage to scale to higher dimensions and enable to build a generic classifier that can capture non-linear tendency to solve the non-stationary issue. However, this method results in numerical instabilities for unbalanced data and are not suited to non-Gaussians likelihoods. That is, the expected improvement function may include fewer candidates for the under-presented functions, yields to miss-representation. To address these issue, we expose new algorithms that can solve these limitations. First, we propose a \emph{linearized link-function} to accounts the under-presented class by using a GP surrogate model. This method is based on Laplace’s method and Gauss-Newton approximations to the Hessian. Our method can improve generalization and be useful when validation data is unavailable (e.g., in non-stationary settings) to solve heteroscedastic behaviours. Our experiments demonstrate that our BO by Gauss-Newton approach competes favorably with state-of-the-art black-box optimization algorithms.

\section{Background}
 Let $\mathcal{D}_n:=\{(\x_n,y_n)\}_{n=1}^N$ denotes a set of data point with a smooth function $f:\mathcal{X}\xrightarrow{}\mathbb{R}$ over a bounded input domain $\mathcal{X} \subseteq \mathbb{R}^d$. The observation $y\sim \Normal(f(\x), \sigma^2)$ denotes the noisy observation of the blackbox function $f$ with noise variance $\sigma^2$. The goal is to find a global optimizer of the function $f$ to trade-off between exploration and exploitation during the search process. A surrogate model is introduced to encode the beliefs over the data points by maximizing an acquisition function which queries the batch-points that maximize the information gain and minimize the uncertainty of the model. We generally use the \emph{Expected Improvement} (\cite{Mockus1978}) acquisition function for its tight link to \emph{density estimation}. We define $\tau = \Phi(\gamma)^{-1}$ a threshold with some constant $\gamma$ that denotes an arbitrary quantile of the observed $y$ values. The utility function $I_{\gamma}(\x)$ quantifies the non-negative improvement over $\tau$ such as
 \begin{align*}
     I_{\gamma}= \max(\tau-y,0),
 \end{align*}
 and the EI function determines new query points by maximising expected gain relative to the the posterior predictive distribution $p(\y|\x,\mathcal{D}_n)$ observed so far
 \begin{align*}
     \alpha_{\gamma}(\x,\mathcal{D}_n)=\Ex_{p(\y|\x,\mathcal{D}_n)}\left[I_{\gamma}(\x)\right].
 \end{align*}
Clearly, when $\gamma = 0$ yields to $\tau=\min_n y_n$, this dependency occurs due to the CDF mapping from $\tau$ to $\gamma$. Following \cite{yamada} we define the $\gamma$-\emph{relative density ratio} of two aribitrary function $\ell(\x)$ and $g(\x)$ as follow
\begin{align*}
    r_{\gamma}(\x) = \frac{\ell(\x)}{\gamma \ell(\x)+\left(1-\gamma\right)g(\x)},
\end{align*}
where $\gamma\ell(\x)+\left(1-\gamma\right)g(\x)$ denotes the $\gamma$-\emph{mixture density} with mixing proportion $0\leq\gamma<1$. We recover the original density ratio $r_0(\x)=\ell(\x)/g(\x)$ when $\gamma=0$, in addition we can express the relative density ratio with the the original one such as
\begin{align*}
    r_{\gamma}(\x) = \left(\gamma+ \frac{g(\x)}{\ell(\x)}\left(1-\gamma\right)\right)^{-1}.
\end{align*}
At this stage we need to define the corresponding distributions of $\ell(\x)$ and $g(\x)$ with respect the conditional distribution $p(\x|\y,\mathcal{D}_n)$ such that
\begin{align*}
p(\x|\y,\mathcal{D}_n)= \left\{ \begin{array}{rcl}
\ell(\x)  & \mbox{if} & y<\tau \\
g(\x)  & \mbox{if} & y\geq\tau
\end{array}\right.
\end{align*}
Interestingly, \cite{NIPS2011_86e8f7ab} showed that the EI acquisition function can be expressed as proportional to $\gamma$-\emph{relative density ratio} such as $\alpha_{\gamma}(\x;\mathcal{D}_n)\propto r_{\gamma}(\x)$. Therefore the global optimization problem is reduced to maximizing the $\gamma$-\emph{relative density ratio}
\begin{align*}
    \x_{\star} = \arg \max_{\x\in \mathcal{X}} \alpha_{\gamma}(\x;\mathcal{D}_n)= \arg \max_{\x\in \mathcal{X}} r_{\gamma}(\x),
\end{align*}
this optimization problem can be optimized using a variety of approaches from \emph{density estimation} literature. In this paper we treat the case of Bayesian neural network parametrized by the hyperparameters $\boldsymbol{\theta}$ that consist of the prior and the likelihood precisions specified by a model $\mathcal{M}$. In particular, for a Bayesian neural network, we explicitly include $\boldsymbol{\theta}$ in the acquisition function as
\begin{align*}
     \alpha_{\gamma}(\x,\boldsymbol{\theta},\mathcal{D}_n)=\Ex_{p(\y|\x,\boldsymbol{\theta},\mathcal{D}_n)}\left[I_{\gamma}(\x, \boldsymbol{\theta})\right],
 \end{align*}
where
\begin{align*}
p(\y|\x, \boldsymbol{\theta},\mathcal{D})=\Ex_{p(\boldsymbol{\theta}|\mathcal{D})}\left[p(\y|\x,\boldsymbol{\theta})\right].    
\end{align*}
A Bayesian model can then be defined using a likelihood and a prior to be able to compute the posterior distribution $p(\boldsymbol{\theta}| \mathcal{D}, \mathcal{M})\propto p(\mathcal{D}|\mathcal{M},\boldsymbol{\theta})p(\boldsymbol{\theta},\mathcal{M})$. In most case this posterior is intractable we then resort to approximation techniques. The main contribution of this work is to link Bayesian optimization to approximate inference using density ratio-estimation.

\section{Methodology}
\subsection{Approximate Acquisition Function for Bayesian Neural Network}

We consider the context of Bayesian neural network (BNN) and our goal is to find an approximate acquisition function that will enable to perform the optimisation loop borrowed from deep learning theory. We first introduce a feature extractor $\f(\x;\boldsymbol{\theta})$ with parameter $\boldsymbol{\theta}\in \mathbb{R}^P$ and use a likelihood function $p(\mathcal{D}|\boldsymbol{\theta})$ that map the inputs to the output class targets $\y \in \mathbb{R}^{C}$, such as 
\begin{align*}
    p(\mathcal{D}|\boldsymbol{\theta})=\prod_{n=1}^N p(\y|\f(\x_n;\boldsymbol{\theta})).
\end{align*}
Generally, in approximate inference theory, we use an inverse link function $\g^{-1}$ such that $\Ex[\y]:=\g^{-1}(\f(\x, \boldsymbol{\theta}))$. By imposing a prior $p(\boldsymbol{\theta})$ on the likelihood parameters we aim to compute the posterior given the data $p(\boldsymbol{\theta}|\mathcal{D})$. Typically, we choose a Gaussian prior $p(\boldsymbol{\theta})=\Normal(\m_0,\S_0)$, we then make a probabilistic prediction for new inputs $\x_{\star}$ using the posterior predictive 
\begin{align*}
p(\y_{\star}|\x_{\star}, \mathcal{D})=\Ex_{p(\boldsymbol{\theta}|\mathcal{D})}\left[p(\y_{\star}|\f(\x_{\star},\boldsymbol{\theta})\right].    
\end{align*}
The exact posterior inference is in general intractable or complicated to compute due to high-dimensional integral, the \emph{model evidence} or \emph{marginal likelihood} $p(\y|\x) = \int p(\mathcal{D}|\boldsymbol{\theta})p(\boldsymbol{\theta})d\boldsymbol{\theta}$ is infeasible in general. We need to resort to approximate inference techniques such as \emph{mean-field-variational-inference} or \emph{Laplace approximation} to approximate $q(\boldsymbol{\theta}) \approx p(\boldsymbol{\theta}|\mathcal{D})=\frac{p(\y|\x,\boldsymbol{\theta})p(\boldsymbol{\theta})}{p(\y|\boldsymbol{\theta})}$. Following \cite{der_two_sample} we can discriminate between two pairs of distributions inputs $\psi_1(\x)$ and $\psi_2(\x)$ and infer the global mixture distribution, therefore we need to express the conditional distribution $p(\x|\y, \mathcal{D}_n)$ in terms of our function value $\f(\x,\boldsymbol{\theta})$. We introduce a linearized form of the neural network using a first-order Taylor approximation around $\boldsymbol{\theta}_{\star}$
\begin{align*}
p(\x|\y,\boldsymbol{\theta},\mathcal{D}_n)= \left\{ \begin{array}{rcl}
\psi_1(\x,\boldsymbol{\theta}):= \f(\x;\boldsymbol{\theta})+\mathcal{J}_{\boldsymbol{\theta}}(\boldsymbol{\theta}-\boldsymbol{\theta}_{\star}) & \mbox{if} & y<\tau \\
\psi_2(\x,\boldsymbol{\theta}):= \f(\x;\boldsymbol{\theta})+\mathcal{J}_{\boldsymbol{\theta}}(\boldsymbol{\theta}-\boldsymbol{\theta}_{\star})  & \mbox{if} & y\geq\tau
\end{array}\right\}
\end{align*}
where $\boldsymbol{\theta}_{\star}=\boldsymbol{\theta}_{\text{map}}$ which corresponds to the \emph{maximum a posteriori}. Note that $\f(\x;\boldsymbol{\theta})$ is not restricted to Gaussian family and can depend non-linearly on $\boldsymbol{\theta}$. We reformulate the EI function to include the posterior distribution $p(\boldsymbol{\theta}|\mathcal{D})$
\begin{align}
    \alpha_{\gamma}(\x;\boldsymbol{\theta}, \mathcal{D}) &= \Ex_{p(\y|\x,\boldsymbol{\theta}, \mathcal{D})}\left[U(\x,\y,\tau)\right]\notag\\
    &=\int_{-\infty}^{\tau}\Ex_{p(\boldsymbol{\theta}|\mathcal{D})}\left[p(\y|\f(\x, \boldsymbol{\theta}))\right](\tau-\y)d\y\notag\\
    &\propto \Ex_{p(\boldsymbol{\theta}|\mathcal{D})}\left[\left(\gamma+(1-\gamma)\frac{\psi_1(\x)}{\psi_2(\x)}\right)^{-1}\right]:=r_{\gamma}(\x,\boldsymbol{\theta},\mathcal{D}),
    \label{eq:full_bayes_ei}
\end{align}
where we follow \cite{yamada} to link the acquisition function to the $\gamma$-relative density ratio, the acquisition function is equivalent to $\gamma$-relative density ratio up to a certain constant factor (\cite{sugiyama_suzuki_kanamori_2012}). Doing so, we recover the formulation given by \cite{bengio_bergstra_algoHPO} under the distribution $p(\boldsymbol{\theta}|\mathcal{D})$. Refer to \hyperref[sec:Appendix.A]{Appendix.A} for the full derivation. The last expression in equation (\ref{eq:full_bayes_ei}) is reduced to an expectation of mixture density with respect to the prior distribution $p(\boldsymbol{\theta}|\mathcal{D})$ with mixing proportion $0\leq\gamma<1$. Since this formulation  is based on \emph{Tree-structured Parzen estimator} (TPE) it provides two advantages: First, we identify a monotonically non-decreasing function which allows to maximize this optimization problem easily. Second, we can estimate separately both $\psi_1(\x)$ and $\psi_2(\x)$ using a tree-based variant of kernel density estimation (KDE) (\cite{Silverman86}). The main advantage of this method is to reduce the computational cost to $\mathcal{O}(N)$ compared to more generic surrogate model as Gaussian processes (GP) which scale to $\mathcal{O}(N^3)$. One major issue is that the $\gamma$-relative density ratio based methods are exacerbated by \emph{error sensitivity} (\cite{variable_kde}), particularly in high dimension and make the optimization problem difficult to solve. To overcome these pitfalls, one alternate solution is to link the approximate acquisition function to \emph{Class-probability-estimation} (CPE) (\cite{der_two_sample}, \cite{cpe_density_book_sugiyama_suzuki_kanamori_2012}, \cite{cpe1}). This method belongs to classification family algorithm and enables to have a closed form of the objective function. First, we follow \cite{linking_cpe_pmlr-v48-menon16} formalism to introduce a binary target variable $z$ to discriminate against the quantile value $\tau$, more formally we have
\begin{align*}
z= \left\{ \begin{array}{rcl}
1 & \mbox{if} & y<\tau \\
0  & \mbox{if} & y\geq\tau
\end{array}\right\}.
\end{align*}
Having defined $\psi_1(\x)=p(\x|z=1, \boldsymbol{\theta})$ and $\psi_2(\x) = 1-\psi(\x|z=1, \boldsymbol{\theta})$, we can apply the Bayes' rule and rewrite the $\gamma$-relative-density function as
 \begin{align*}
    r_{\gamma}(\x,\boldsymbol{\theta},\z,\mathcal{D})&=\Ex_{p(\boldsymbol{\theta}|\mathcal{D})}\left[\frac{\psi_1(\x,\boldsymbol{\theta})}{\gamma \psi_1(\x,\boldsymbol{\theta})+(1-\gamma)\psi_2(\x,\boldsymbol{\theta})}\right]\\
    &=\gamma^{-1}\Ex_{p(\boldsymbol{\theta}|\mathcal{D})}\left[\pi(\x,\boldsymbol{\theta})\right],
 \end{align*}
 where we established the link between the class-probability and the $\gamma$-relative density ratio. In comparison to TPE, this method is always bounded above $1/\gamma$ for all $\gamma>0$ and does not suffer from singularity. In the case where we approximate the posterior by a variational distribution $q(\boldsymbol{\theta})\approx p(\boldsymbol{\theta}|\mathcal{D})$ we obtain an approximate expression denoted by $\hat{r}_{\gamma}(\x;\boldsymbol{\theta}, \mathcal{D})$. Refer to the \hyperref[sec:Appendix.B]{Appendix.B} for the full derivation. Ultimately, in Bayesian inference, we wish to maximize the \emph{marginal} acquisition function which marginalizes out the uncertainty about the hyperparameters.
 \subsection{Model Evidence}
 In this Bayesian learning framework, the ultimate goal is to maximize the \emph{marginal likelihood} function which depends on the acquisition function under the variational distribution $q(\boldsymbol{\theta})$. Now we turn our attention to express the likelihood function $p(\z|\x,\boldsymbol{\theta})=\text{Bernoulli} (\z|\pi(\x;\boldsymbol{\theta}))$ in order to formulate the optimization problem. We give an explicit formulation about how the acquisition function depends on the hyperparameters wrt the variational distribution $q(\boldsymbol{\theta})$ and how it should be optimized. That is, we specify a prior $p(\boldsymbol{\theta})$ for a particular model $\mathcal{M}$, the model might consists of the choice of network architecture, hyperparameters of the likelihood and prior variance. We are interested in maximizing the marginal class-posterior probabilities considered as the objective function, the optimization problem takes the following form
\begin{align*}
    \arg \max _{\boldsymbol{\theta}}\Pi(\x,\mathcal{D}|\mathcal{M}),
\end{align*}
optimizing the posterior class-probabilities is equivalent to optimizing the marginal acquisition function, we have
\begin{align}
    \Pi(\x,\mathcal{D}|\mathcal{M})&=\int p(\z|\x,\boldsymbol{\theta})p(\boldsymbol{\theta}|\mathcal{D})d\boldsymbol{\theta}\notag\\
    &\approx\frac{1}{q(\mathcal{D}|\mathcal{M})}\biggl(\gamma\Ex_{q(\boldsymbol{\theta})}\left[\pi(\x;\boldsymbol{\theta}_{\star})\right]+(1-\gamma)\Ex_{q(\boldsymbol{\theta})}\left[1-\pi(\x;\boldsymbol{\theta}_{\star})\right])\biggr)\label{eq:Pi},
\end{align}
where $q(\mathcal{D}|\mathcal{M})$ denotes the approximate marginal likelihood for target values $\y$. The expression above in equation (\ref{eq:Pi}) uses the marginal likelihood for neural network in order to approximate the posterior distribution (\cite{mackay1992a}). This method is based on a local quadratic approximation of $p(\boldsymbol{\theta}|\mathcal{D})$ around the maximum $\boldsymbol{\theta}_{\star}$ resulting in a Gaussian approximation denoted by $q(\boldsymbol{\theta})$ and an approximation to the marginal likelihood denoted by $q(\mathcal{D}|\mathcal{M})$ given the model $\mathcal{M}$. In particular, we have
\begin{align}
    \log p(\mathcal{D}|\mathcal{M})\approx \log q(\mathcal{D}|\mathcal{M}):
    &=\log p(\mathcal{D},\boldsymbol{\theta}_{\star}|\mathcal{M})-\left(\boldsymbol{\theta}-\boldsymbol{\theta}_{\star}\right)^T\mathcal{J}_{\boldsymbol{\theta}_{\star}}-\frac{1}{2}\left(\boldsymbol{\theta}-\boldsymbol{\theta}_{\star}\right)^T\mathcal{H}_{\boldsymbol{\theta}_{\star}}\left(\boldsymbol{\theta}-\boldsymbol{\theta}_{\star}\right)^T\notag\\
    &=\log p(\mathcal{D}, \boldsymbol{\theta}_{\star}|\mathcal{M})-\frac{1}{2}\log \text{det}\left(\frac{1}{2\pi}\mathcal{H}_{\boldsymbol{\theta}_{\star}}\right)\label{eq:marginal_lik}
\end{align}
where $\mathcal{H}:=-\nabla_{\boldsymbol{\theta \theta}}^{2}\log p(\mathcal{D},\boldsymbol{\theta}_{\star})$ denotes the Hessian matrix and $\mathcal{J}:=-\nabla_{\boldsymbol{\theta}}\log p(\mathcal{D},\boldsymbol{\theta}_{\star})$ denotes the gradient evaluated at $\boldsymbol{\theta}_{\star}$. Exponentiating the approximation to the joint distribution in equation (\ref{eq:marginal_lik}) and dividing by the marginal likelihood, we obtain the Laplace approximation as a Gaussian distribution with mean $\boldsymbol{\mu}=\boldsymbol{\theta}_{\text{map}}$ and covariance $\boldsymbol{\Sigma}=-\left[\nabla_{\boldsymbol{\theta}\boldsymbol{\theta}}^2\log p(\mathcal{D},\boldsymbol{\theta})\rvert_{\boldsymbol{\theta}=\boldsymbol{\theta}_{\text{\emph{map}}}}\right]^{-1}$.
The full derivation is given in \hyperref[sec:Appendix.D]{Appendix.D}. This expression shows that the optimization of the surrogate model is reduced to maximizing the approximate $\gamma$-relative-density ratio $\hat{r}_{\gamma}(\x,\boldsymbol{\theta}, \z,\mathcal{D})$. We can express the objective function as the \emph{proper scoring rule}, we are therefore interested to minimize 
\begin{align}
    \mathcal{L}(\boldsymbol{\theta}|\mathcal{D},\gamma,\mathcal{M}):=-\frac{1}{N}\biggl(\sum_{n=1}^N\log \pi(\x_n,\boldsymbol{\theta})\gamma+(1-\pi(\x_n,\boldsymbol{\theta}))(1-\gamma)
    -\sum_{n=1}^N\log q(\mathcal{D}_n|\mathcal{M})\biggr)\label{eq:final_objective}.
\end{align}
The last expression in equation (\ref{eq:final_objective}) shows that we would like points $\x$ with high probability under the posterior-class probability $\pi(\x;\boldsymbol{\theta})$ and low probability under $1-\pi(\x;\boldsymbol{\theta})$. Since we use the tree-structured formalism, we first sample from the approximate distribution $q(\boldsymbol{\theta})$ and evaluate the expected ratio according to class-binary variable $\z$. On each iteration, the algorithm returns the candidate $\x_{\star}$ with the greatest expected improvement. See (Algorithm\ref{algo:algo_1}) for the pseudo-code of the proposed method.

\subsection{The choice of the approximate distribution}
\label{sec:choice_q_approx}
There is mainly two concurrent approaches in approximate inference: i) \emph{Mean-field Variational Inference (VI)}: this method approximates the posterior $p(\boldsymbol{\theta}|\mathcal{D})$ by a decoupled variational distribution $q(\boldsymbol{\theta})$ optimized by maximizing the \emph{evidence-lower-bound} or ELBO to the marginal likelihood (\cite{blundell2015weight}). ii) \emph{Laplace approximation}: proposed by \cite{mackay1992a} is based on approximating the posterior by a Gaussian distribution around the mode $\boldsymbol{\theta}_{\text{\emph{map}}}$ and covariance matrix $\boldsymbol{\Sigma}$ with $q(\boldsymbol{\theta}_{\text{\emph{map}}}, \boldsymbol{\Sigma})$
\begin{align*}
    \boldsymbol{\Sigma}=-\left[\nabla_{\boldsymbol{\theta}\boldsymbol{\theta}}^2\log p(\mathcal{D},\boldsymbol{\theta})\rvert_{\boldsymbol{\theta}=\boldsymbol{\theta}_{\text{\emph{map}}}}\right]^{-1}.
\end{align*}
We follow the approach of Laplace approximation, to compute the covariance, we need to compute the Hessian of $p(\boldsymbol{\theta}|\mathcal{D})$ wrt $\boldsymbol{\theta}$.
But computing the Hessian involves computing the Jacobian $\mathcal{J}^{f}\in \mathbb{R}^{C\times P}$ and the Hessian $\mathcal{H}^{f} \in \mathbb{R}^{C\times P\times P}$ of the function value $\f(\x,\boldsymbol{\theta})$, more formally we have
\begin{align*}
    \nabla_{\boldsymbol{\theta}}\log p(\y|\f(\x, \boldsymbol{\theta})) &= \mathcal{J}^{f}(\x)^T\nabla_{\boldsymbol{\f}}\log p(\y|\f)\\
    \nabla_{\boldsymbol{\theta\theta}}^2\log p(\y|\f(\x, \boldsymbol{\theta})) &= \mathcal{H}^{f}_{\boldsymbol{\theta}}(\x)^T\nabla_{\boldsymbol{\f}}\log p(\y|\f)-\mathcal{J}^{f}_{\boldsymbol{\theta}}(\x)^T\nabla_{\boldsymbol{\f\f}}^2\log p(\y|\f)\mathcal{J}_{\boldsymbol{\theta}}(\x),
\end{align*}
where 
\begin{align*}
    \mathcal{H}^{f}_{\boldsymbol{\theta}}(\x) = \frac{\partial^2 f_c(\x,\boldsymbol{\theta})}{\partial \theta_i \partial\theta_j}\quad \text{ and }\mathcal{J}^{f}_{\boldsymbol{\theta}}(\x) = \frac{\partial f_c(\x,\boldsymbol{\theta})}{\partial \theta_i},
\end{align*}
when dealing with network Hessian, this becomes infeasible and we resort to approximate the covariance, we can therefore approximate $\mathcal{H}_{\boldsymbol{\theta}}(\x)$ using \emph{generalized Gauss-Newton} (GGN) approximation as in \cite{martens_hessian} and \cite{hessian_2002}, then we can rewrite the approximate covariance as
\begin{align*}
    \nabla_{\boldsymbol{\theta\theta}}^2\log p(\y|\f(\x, \boldsymbol{\theta}))\approx -\mathcal{J}^{f}_{\boldsymbol{\theta}}(\x)^T\nabla_{\boldsymbol{\f\f}}^2\log p(\y|\f)\mathcal{J}^{f}_{\boldsymbol{\theta}}(\x).
\end{align*}
This approximation guarantees to be positive semi-definite and can be further approximated to diagonal and Kronecker-factored approximation for efficient storage and inverse computation. In particular, the last expression assumes that $\mathcal{H}^{f}_{\boldsymbol{\theta}}(\x)^T\nabla_{\boldsymbol{\f}}\log p(\y|\f)=0$ (\citet{bottou2018optimization}). Since we have defined the function value  $\f(\x;\boldsymbol{\theta})$ as a linear function of its predictor, yields the Hessian to vanish. Further this linearization will allow to convert the original function value expression to a \emph{generalized linear model} (GLM) which enable to perform space function inference.   
\begin{algorithm}
    \caption{Bayesian optimization by Gauss Newton density estimation (BOGGN)}\label{algo:algo_1}
    \begin{algorithmic}
        \STATE \textbf{Input: }
        Blackbox $f:\mathcal{X}\xrightarrow{}\mathbb{R}$, proportion $\gamma \in (0,1)$, $\pi(\x,\boldsymbol{\theta}):\mathcal{X}\xrightarrow{}[0,1]$
            \WHILE{under budget}
                \STATE  $\tau\xleftarrow{}\Phi^{-1}(\gamma)$ \CommentSty{//compute quantile}
                \STATE $z_n\xleftarrow{}\I[\y_n\leq \tau]_n^N$ \CommentSty{//assign labels}
                \STATE  $\Tilde{\mathcal{D}}_N\xleftarrow{}\{\x_n,z_n\}_n^N$\CommentSty{//construct auxilliary dataset}
                \STATE $p(\boldsymbol{\theta})\xleftarrow{}\Normal(0,\I\delta^{-1})$\CommentSty{//choose approximate distribution prior} 
                \STATE  $\boldsymbol{\theta}_{\star}\xleftarrow{}\arg \min_{\boldsymbol{\theta}}\mathcal{L}(\boldsymbol{\theta}|\mathcal{D},\gamma,\mathcal{M})$ \CommentSty{//Train the classifier} 
                \STATE  $q(\boldsymbol{\theta})\xleftarrow{}\Normal(\boldsymbol{\theta}_{\star}=\boldsymbol{\theta}_{\text{\map}}, \boldsymbol{\Sigma}_{ggn}^{-1})$ \CommentSty{//compute posterior means and covariance under GGN}
                \STATE 
                $\x_N\xleftarrow{}\arg \max_{\x\in \mathcal{X}}\Ex\left[\pi(\x,\boldsymbol{\theta}_{\star})\right]$ \CommentSty{//suggest candidate wrt classifier}
                \STATE 
                $y_N\xleftarrow{}f(\x_N)$\CommentSty{//evaluate Blackbox function}
                \STATE 
                $\mathcal{D}_N\xleftarrow{}\mathcal{D}_{N-1}\cup\{(\x_N, y_N)\}$ \CommentSty{//update dataset}
            \ENDWHILE
    \end{algorithmic}
\end{algorithm}

\section{Related Work}
Bayesian optimization literature has been extensively based on modeling a probabilistic surrogate model which plays an important role to compute the acquisition function. The common choice is a Gaussian process (GP) due to its flexibility and ability to handle uncertainity, however the GP-based BO suffers from scalability issue and specially when dealing with neural networks as in \cite{snoek_BO_DNN}. Specific treatment has been done in this direction as in \cite{white2020bananas} and in \cite{perrone_transfer_lear} for transfer learning. Recently, a body of work has been extended to handle non-stationary data which is the biggest limitation of GPs, \cite{bohamian} used a flexible parametric model as neural network to overcome the scalability issue through stochastic gradient Hamiltonian Monte Carlo. The most common pitfalls of surrogate model is the difficulty to handle discrete and conditional variables, \cite{bengio_bergstra_algoHPO} showed a simplistic technique by applying the density ratio using TPE technique which can overcome this issue by approximating the acquisition function. Expected Improvement (EI) \cite{ei_acq_jones_efficient_1998}
acquisition function remains a good choice, because it is easy to evaluate and to optimize and consistently performs well in practice. A different body of works has been extended to Kernel density estimation (KDE), many alternatives have been proposed in this direction including KL importance estimation procedure (KLIEP) (\cite{cpe_density_book_sugiyama_suzuki_kanamori_2012}),  unconstrained least-squares importance fitting (ULSIF)in \cite{greton_covariate_shift} which based on covariate shift, and relative ULSIF in (\cite{yamada}).  
A relatively closed to our work, \cite{implicit_model_kleinegesse19a} proposed implicit models where the likelihood is intractable but the sampling from the model is possible, they use utility function to link the mutual information between parameters and the data, they achieve this by utilising Likelihood-Free Inference by Ratio Estimation (LFIRE) to approximate posterior distributions. In our work, we make use of density ratio estimation approach and we restrict ourselves to link the CPE (\cite{linking_cpe_pmlr-v48-menon16}, \cite{tiao2021-bore}) using approximate inference techniques. In fact, the main contribution is to formulate a tractable likelihood by a parametrized neural network model yielding to an efficient optimization based on Gauss Netwon methods.     

\section{Experiments}
We describe the experiments conducted to empirically evaluate the efficacy of our method. We empirically evaluate the classifier for the Laplace-GGN approximated in weight space and the corresponding GP classifier in function space. We use a diagonal prior $p(\boldsymbol{\theta})=\Normal(0,\I_P/\delta)$. We first train the network to find the \emph{map} estimate using the objective function in equation (\ref{eq:final_objective}). We then compute the posteriors predictive using the values of the parameters after training $\boldsymbol{\theta}_{\star}$. In a second step, we choose the classifier $\pi(\x,\boldsymbol{\theta})$ as a  multi-layer perceptron (MLP), with 2 hidden layers, each with $32$ units and relu activation function. We optimize the weights with ADAM (\cite{kingma2017adam}) using batch size of $B = 32$. The input-classifier is optimized with respect to its output using multi-started L-BFGS, then a random candidate is applied at a proportion of $\epsilon=0.1$ of the time to encourage exploration. 

\begin{figure}%
    \centering
    {{\includegraphics[scale=0.3]{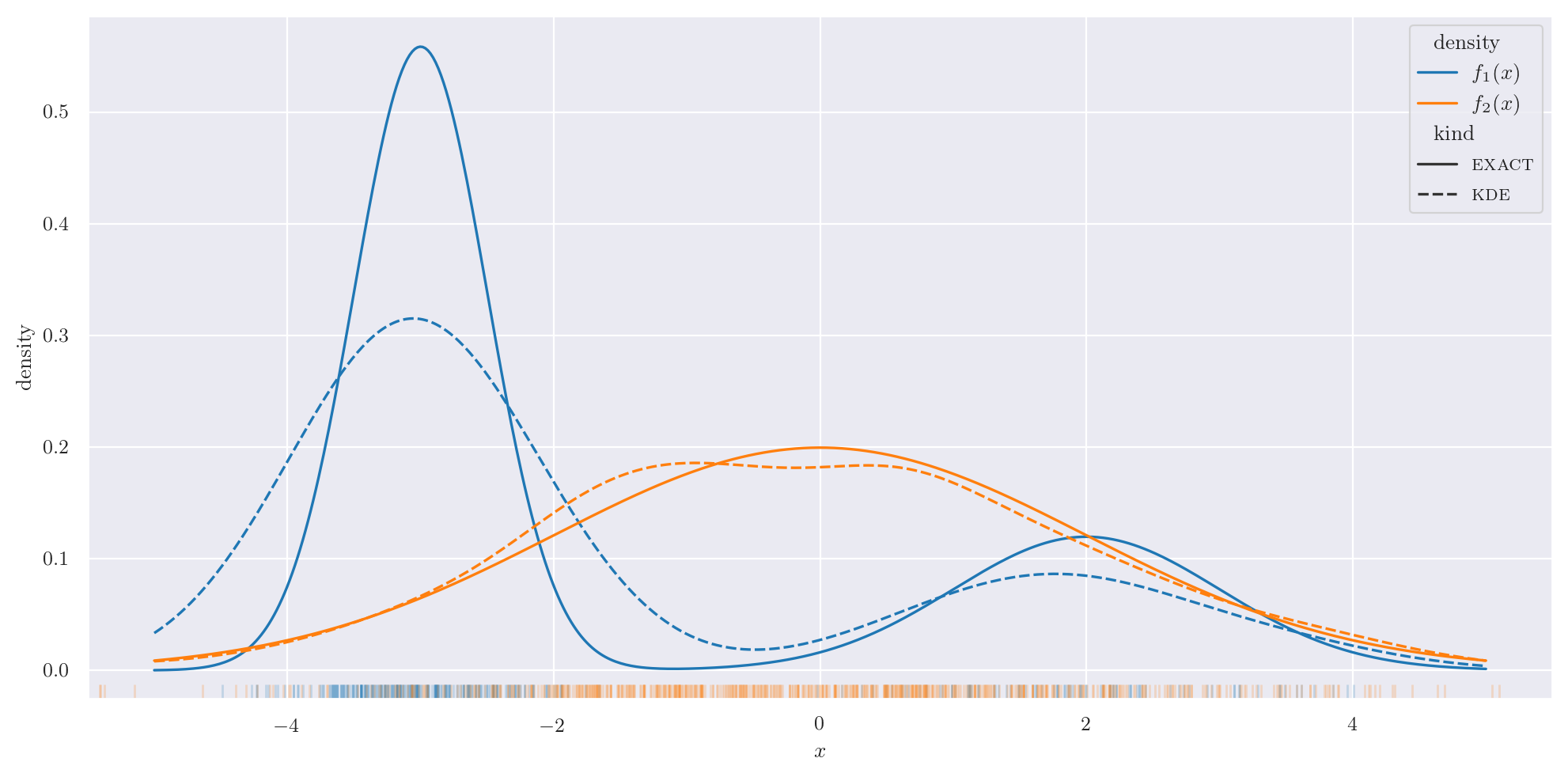}}}\\%
    \centering
    {{\includegraphics[scale=0.25]{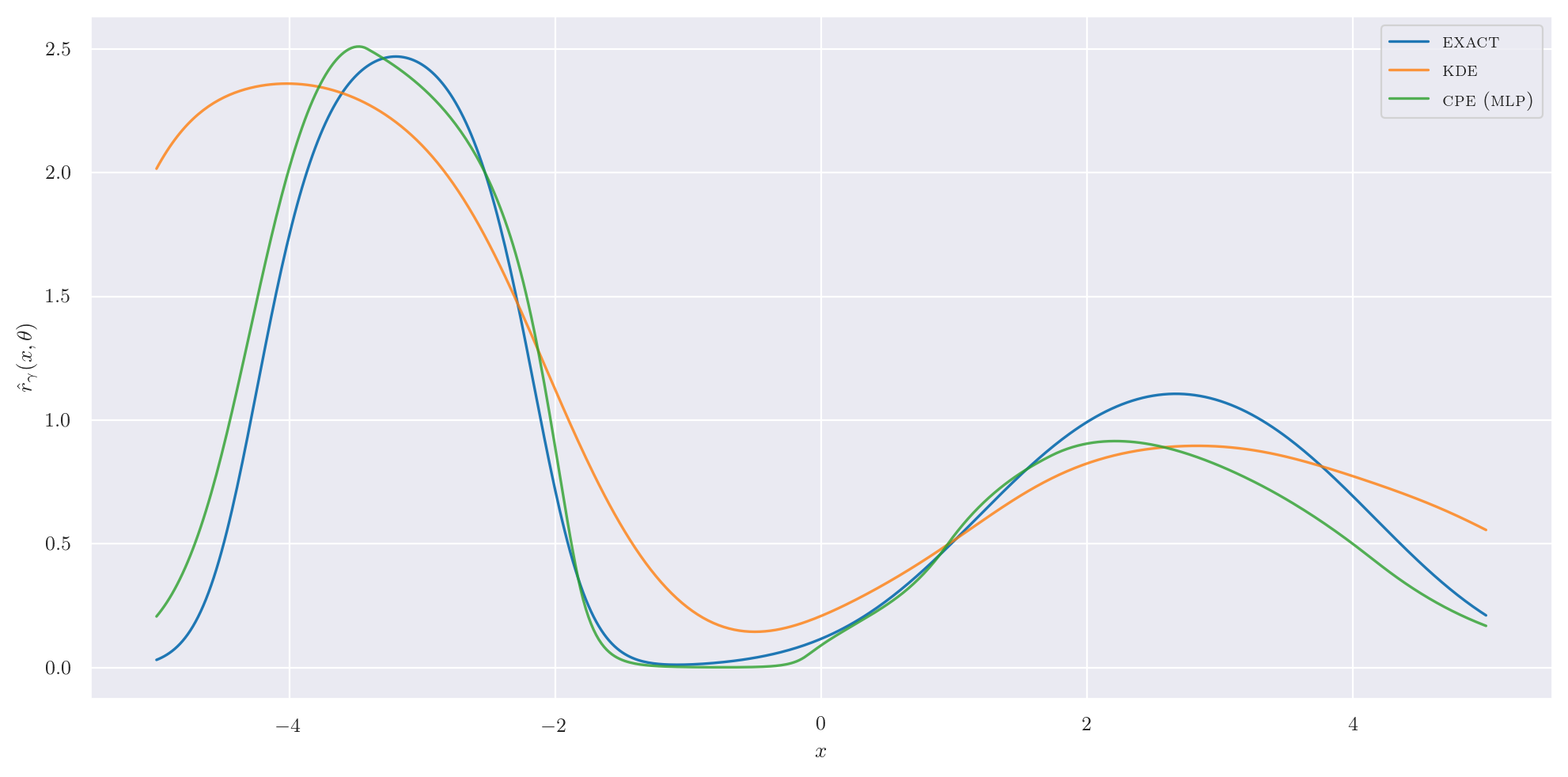}}}%
     {{\includegraphics[scale=0.25]{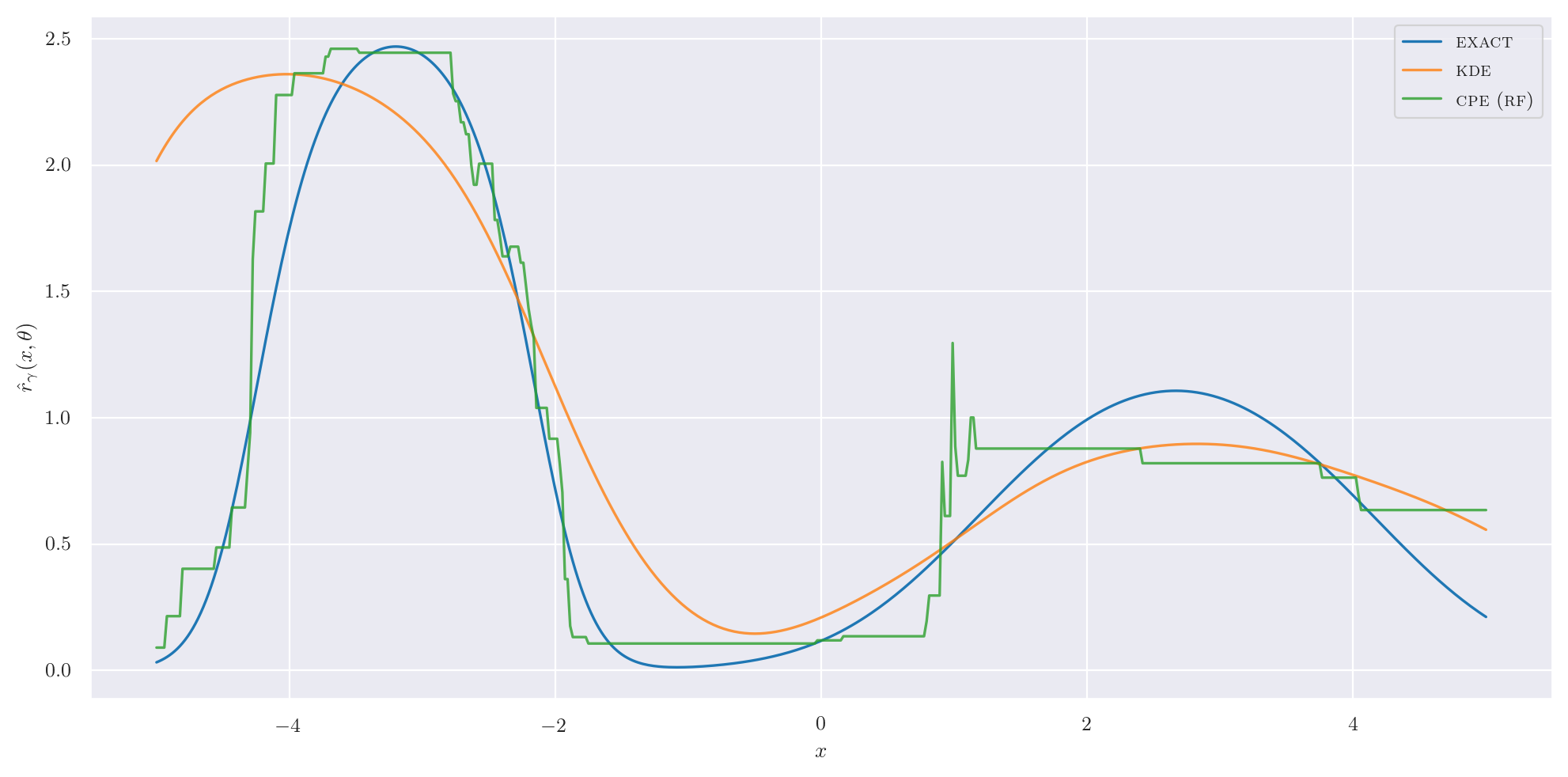}}}\\%
    {{\includegraphics[scale=0.25]{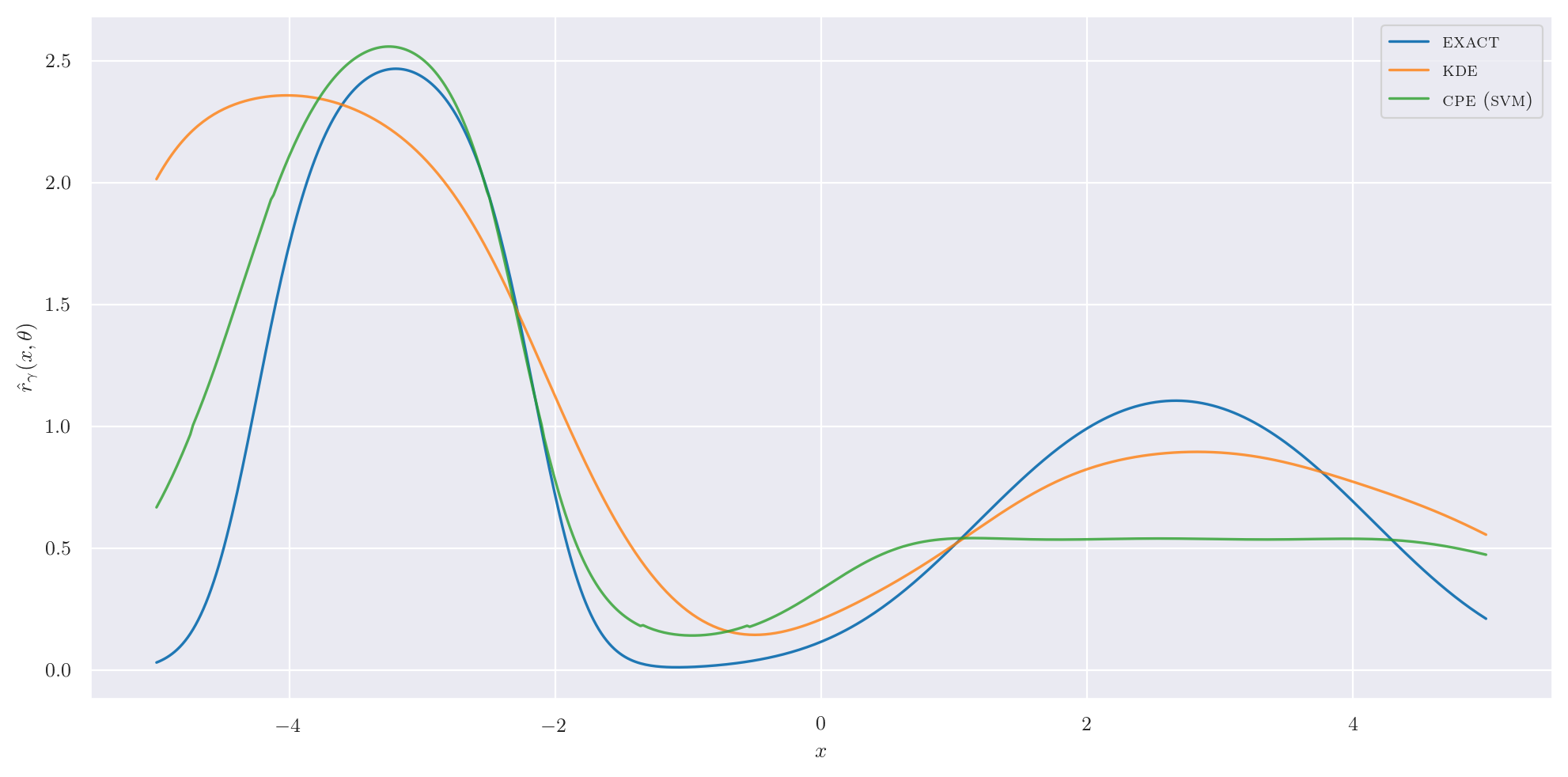}}}%
    {{\includegraphics[scale=0.25]{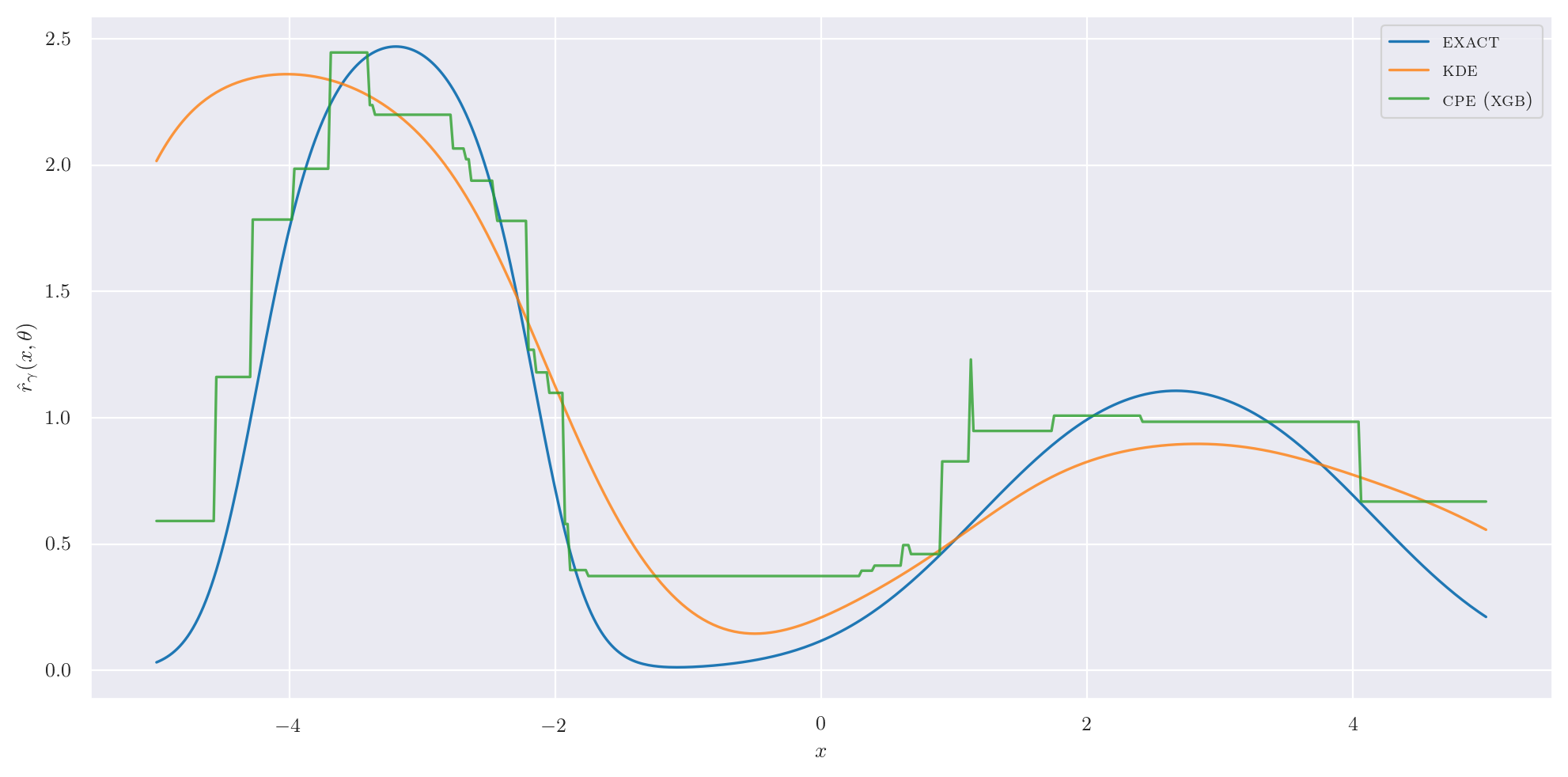}}}
    \caption{Relative density-ratio estimation by probabilistic classification}%
    \label{fig:relative_class}%
\end{figure}
\subsection{Class-posterior probability function}

To represent the true relative density-ratio, we estimate the class-probability estimation (CPE) by the neural network under different variants. In the figure \ref{fig:relative_class}, in top left of the second panel, the probabilistic classifier consists of a simple MLP (multi-layer percepton) with 3 hidden layers, each with and 32 units and elu activations. All other subfigures, we show the same, but with random forest (RF), XGBOOSTand SVM classifiers. The class-posterior probability seems to recover the exact density ratio accurately, whereas the KDE method does so quite poorly. In the context of BO, we are interested to look at the mode of the density-ratio functions. We clearly see that our proposed method are well calibrated. KDE fits $f_1(x)$ well and recovered the modes of $f_2(x)$ accurately however a significant shift in the mode (maximum) of the resulting approximate density-ratio function is caused by an overestimation of the variance of $f_2(x)$.

\subsection{Bayesian Neural network with HPOBench Benchmark}
We first consider a regression problem, by training a feed-forward neural network with two-layer using ADAM optimizer (\cite{kingma2017adam}), the objective function is defined as the validation mean-squared error (MSE). We make use of HPOBench (\cite{klein2019tabular}) which tabultes all combinations of resulting MSEs configurations, the hyperparameters are reportd as : the initial learning rate, learning rate schedule, batch size, widths, activation function type and dropout rates. We consider synthetic and real-world \emph{Protein} datasets.
The figure (\ref{fig:relative_class}) show the results across all selected datasets that our BOGGN method consistly outperforms all other baselines and converge rapidely toward the global minimum in reasonable evaluations.  

We consider generic synthetic test functions well known in Bayesian optimization benchmarks such as \emph{Branin}, \emph{Six-Hump Camel} and \emph{Hart-Mann}. We assess the performance by reporting the \emph{immediate regret} defined as the absolute error between the global minimum and the lowest function value achieved so far. To assess our method performance against BO baselines, We report the \emph{immediate regret} for each benchmark and optimization method with $\gamma = 1/3$ by aggregating across $200$ replicated runs. To account discrete domains, we use Random Search with a function evaluation limit of $500$ for candidate suggestion and $2000$ for continuous domains. We assesses the efficacy of our method against state-of-the-art optimization methods on the different-surrogate-model: \emph{TPE} (\cite{bengio_bergstra_algoHPO}), GP-based BO (GP) (\cite{robo_klein-bayesopt17}), Hyperband (\cite{bohb-icml-18}), Random Search (\cite{bengio_bergstra_algoHPO}) and Bore (\cite{tiao2021-bore}). The figure (\ref{fig:relative_class}) shows clearly that our proposed method BOGGN outperforms all other methods. This can be explained by the fact that the generative family of approximate distribution is able to extract the most useful feature in the domain space, this is done by integrating out all possible uncertainties and averaging over all queries of the domain search. 
\begin{figure}%
    \centering
    {{\includegraphics[scale=0.50]{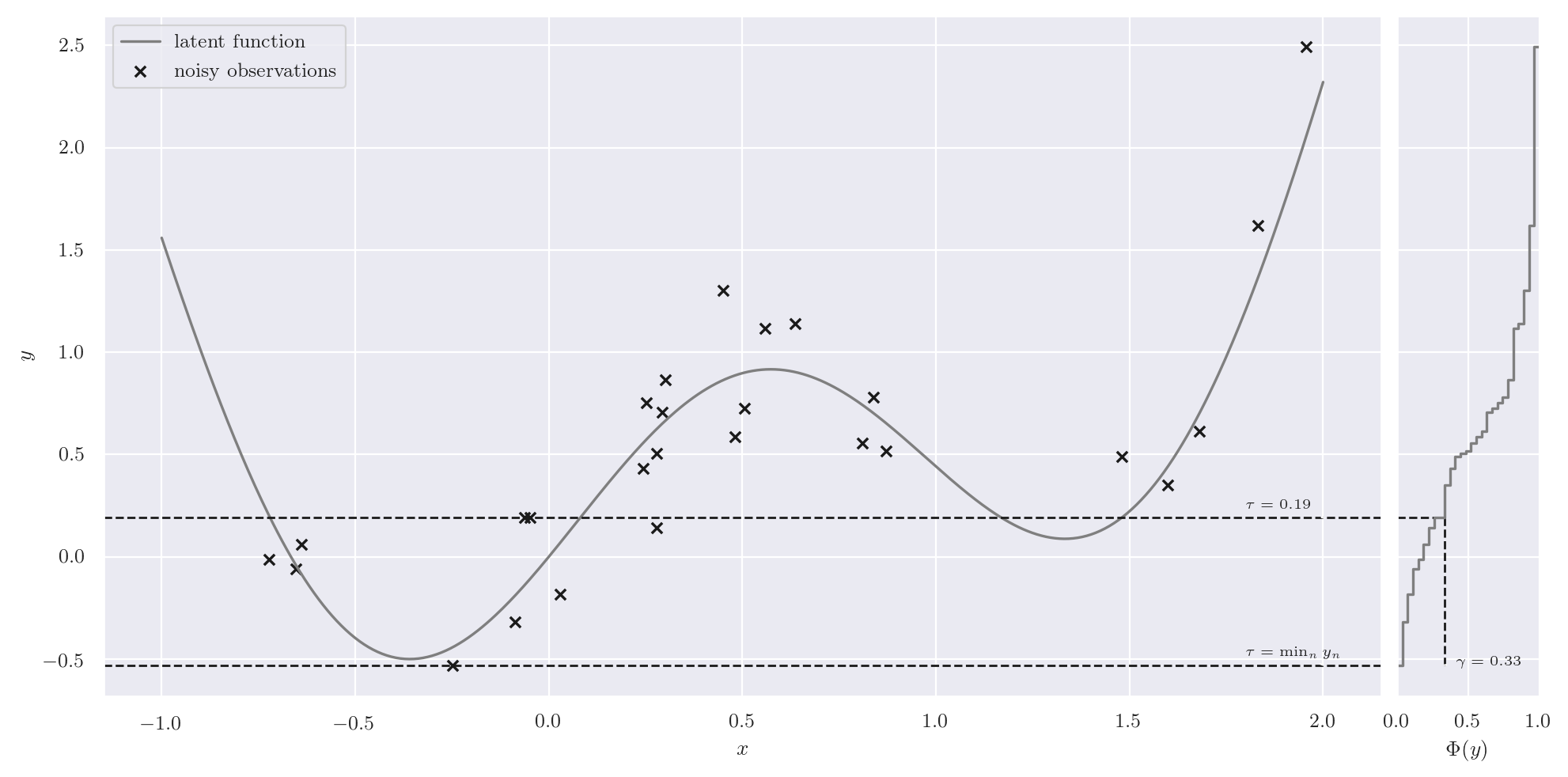}}}%
    \caption{Optimization workflow on a synthetic function}%
    \label{fig:optim_wf}%
\end{figure}


\begin{figure}%
    \centering
    {{\includegraphics[scale=0.25]{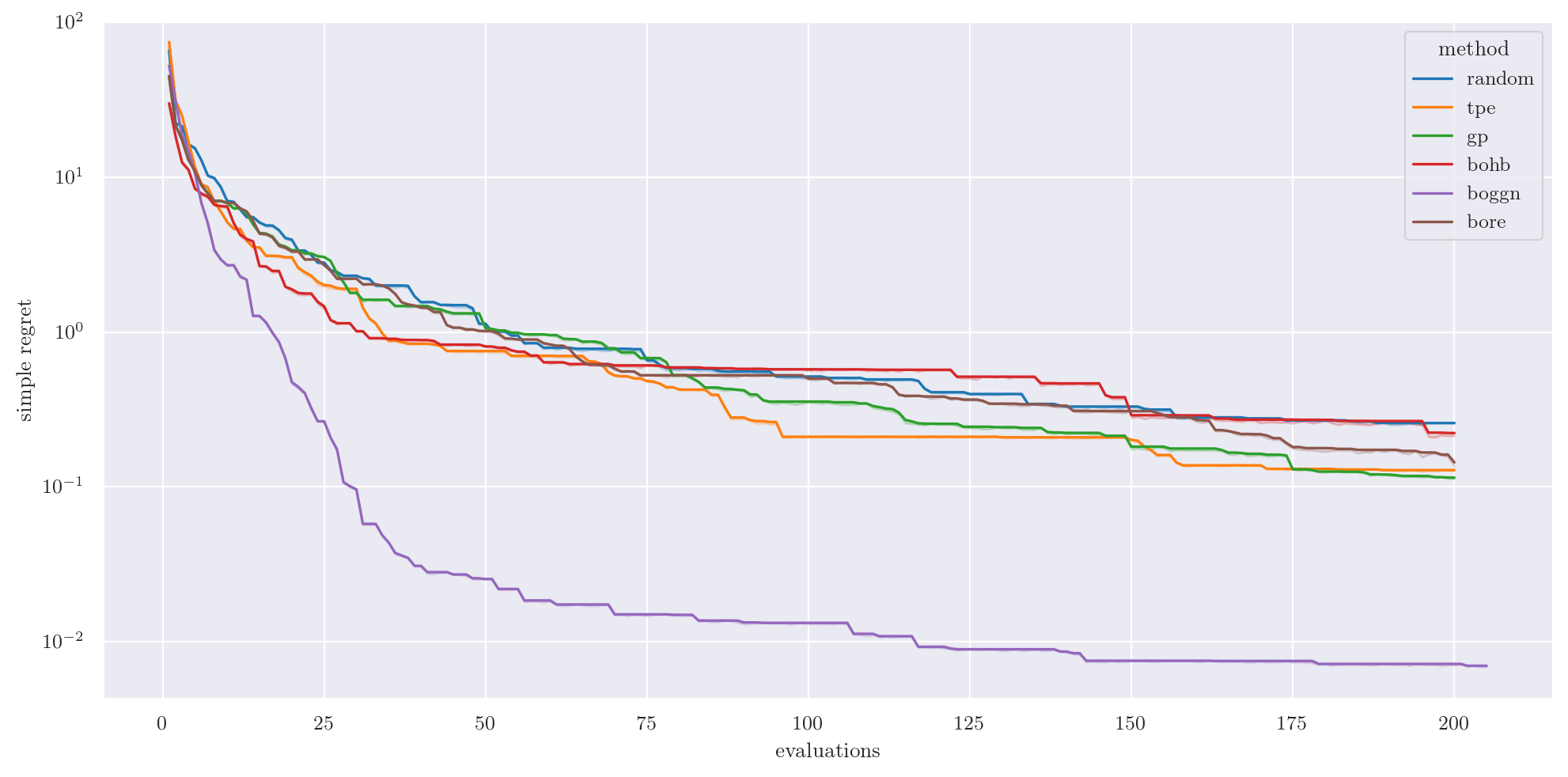}}}%
     {{\includegraphics[scale=0.25]{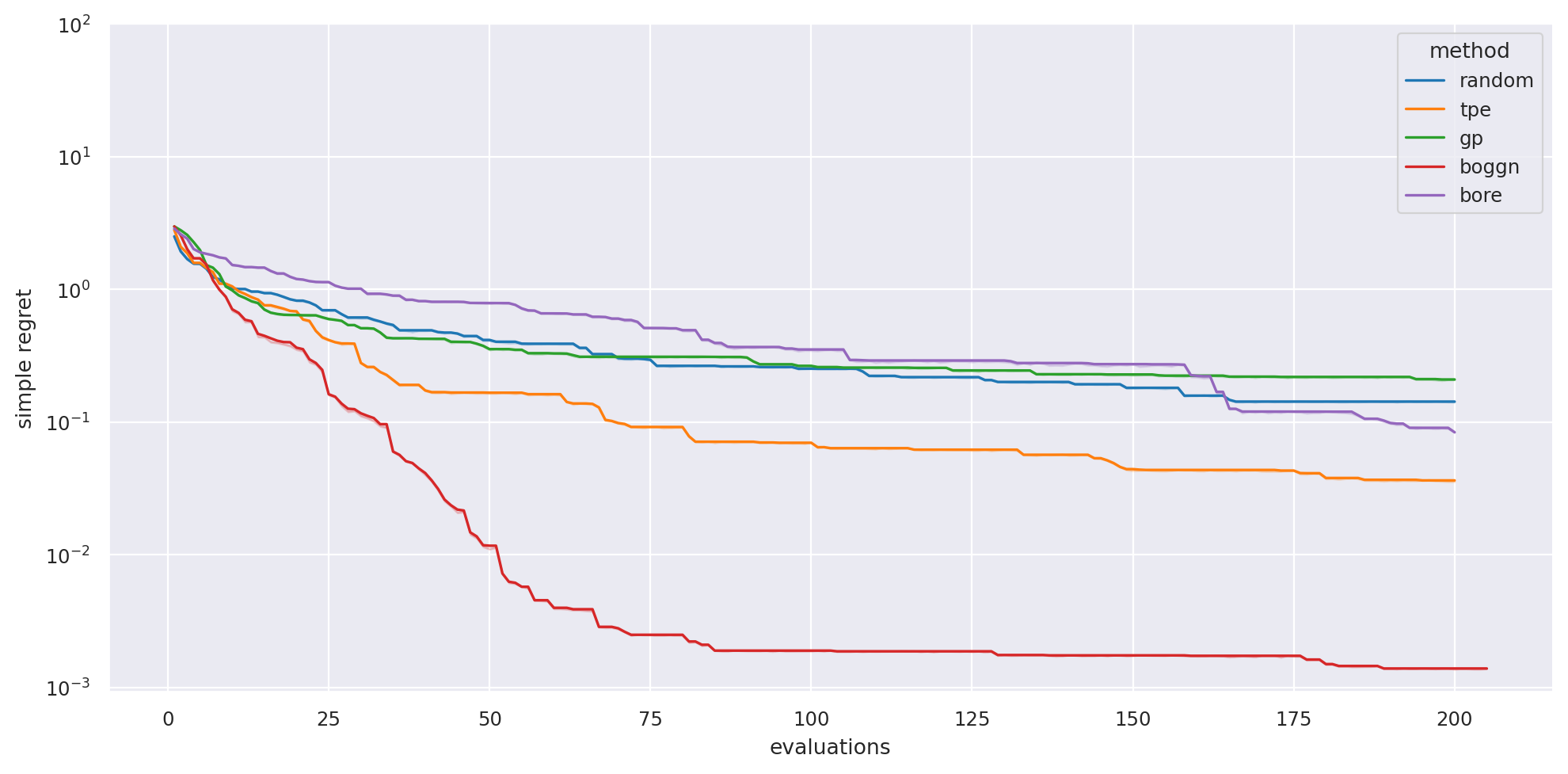}}}\\%
    {{\includegraphics[scale=0.25]{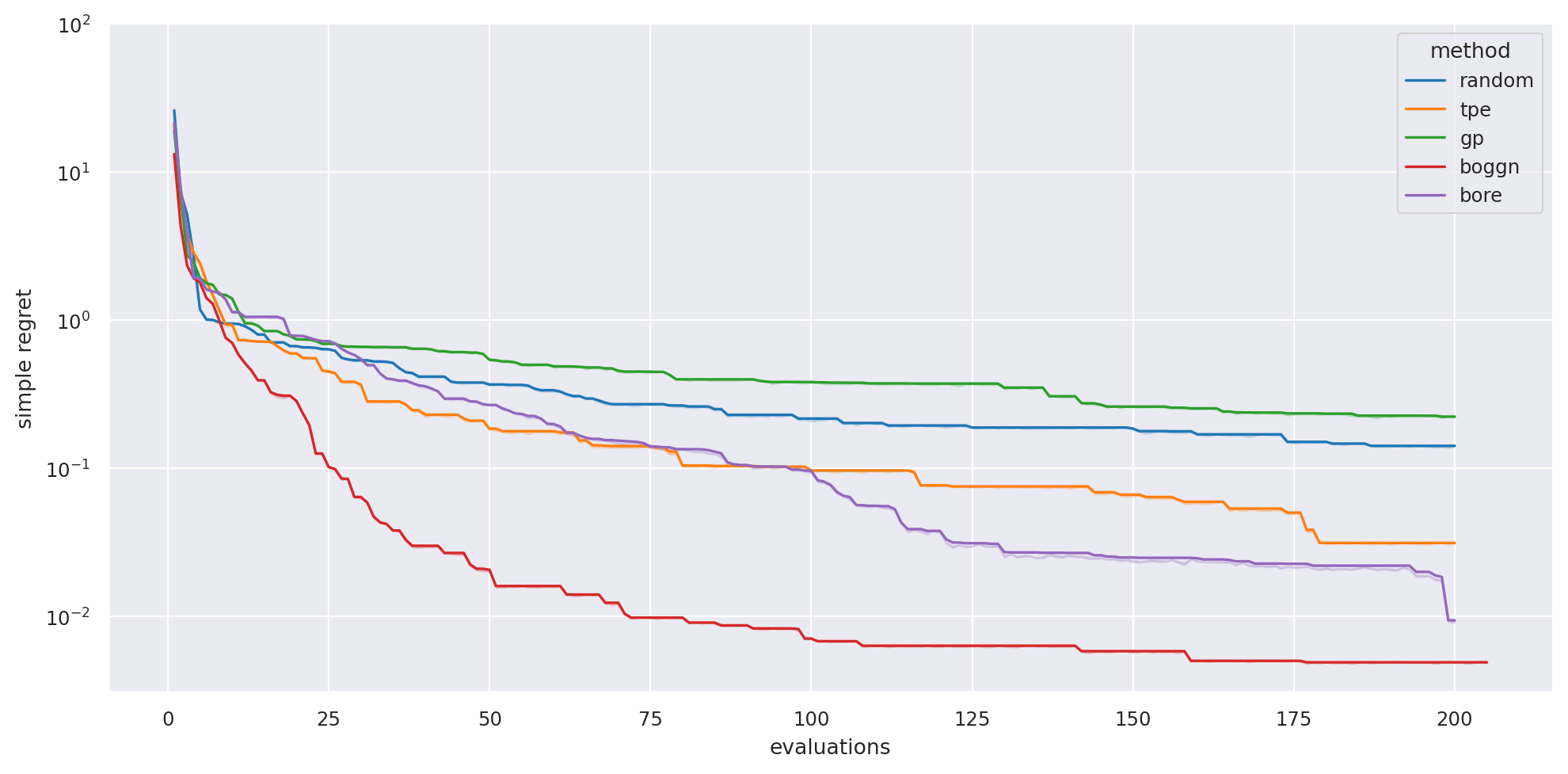}}}
    {{\includegraphics[scale=0.25]{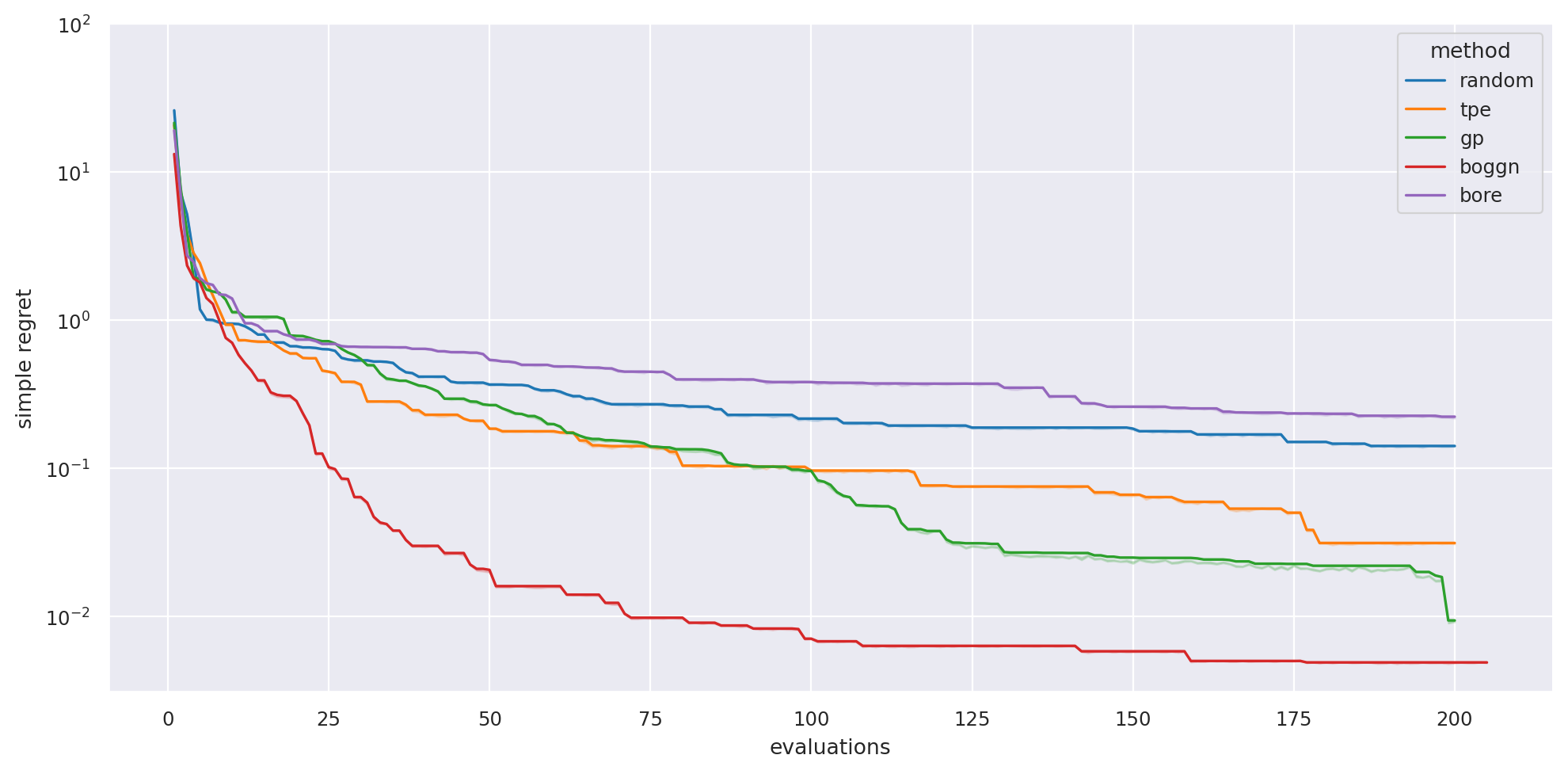}}}
    \caption{Immediate regret over function evaluations on the HPOBench neural network dataset}%
    \label{fig:relative_class_regret}%
\end{figure}

\subsection{Effect of prior on class imbalance}
\begin{figure}[ht]%
    \centering
    {{\includegraphics[scale=0.30]{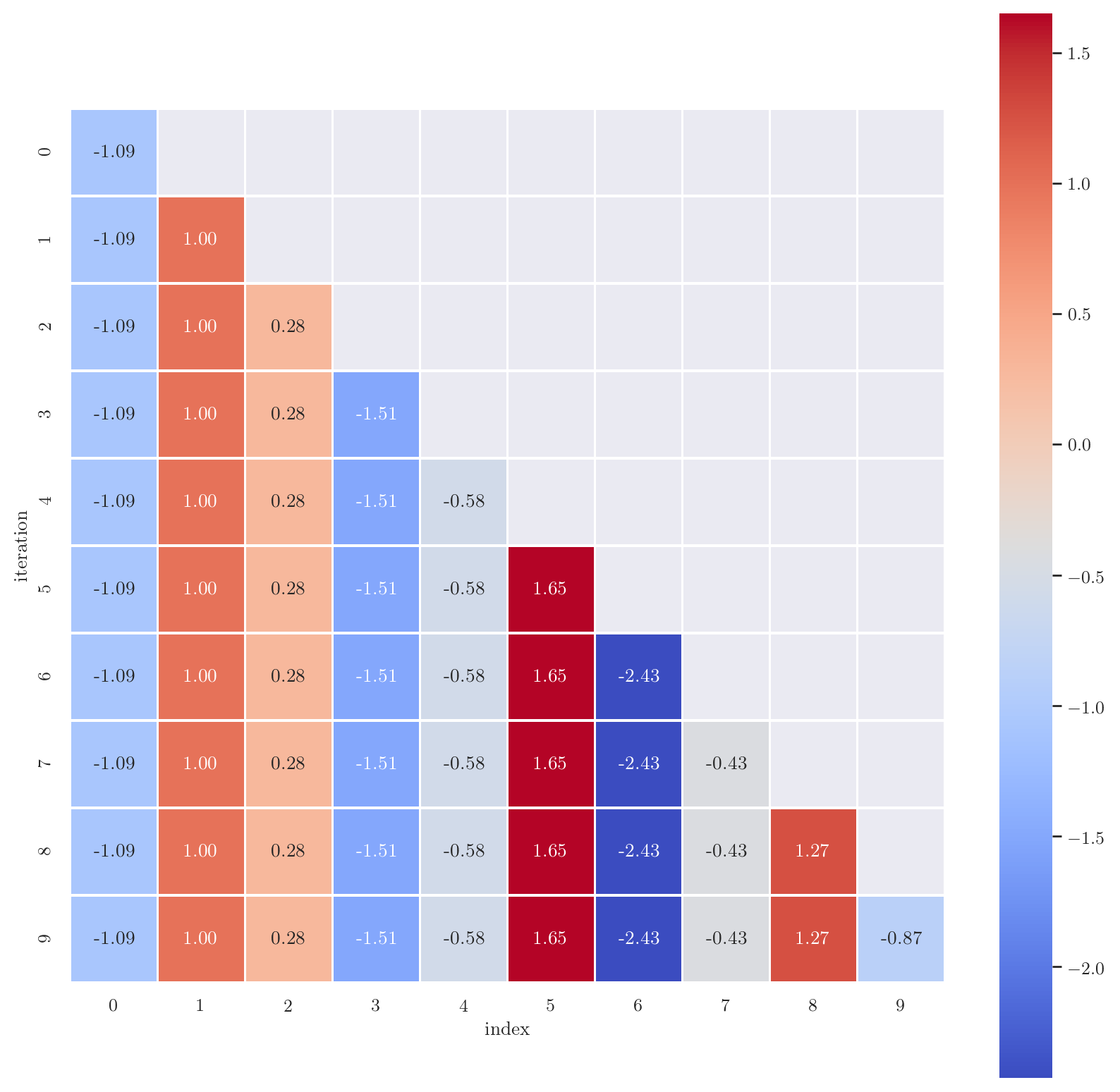}}}%
     {{\includegraphics[scale=0.30]{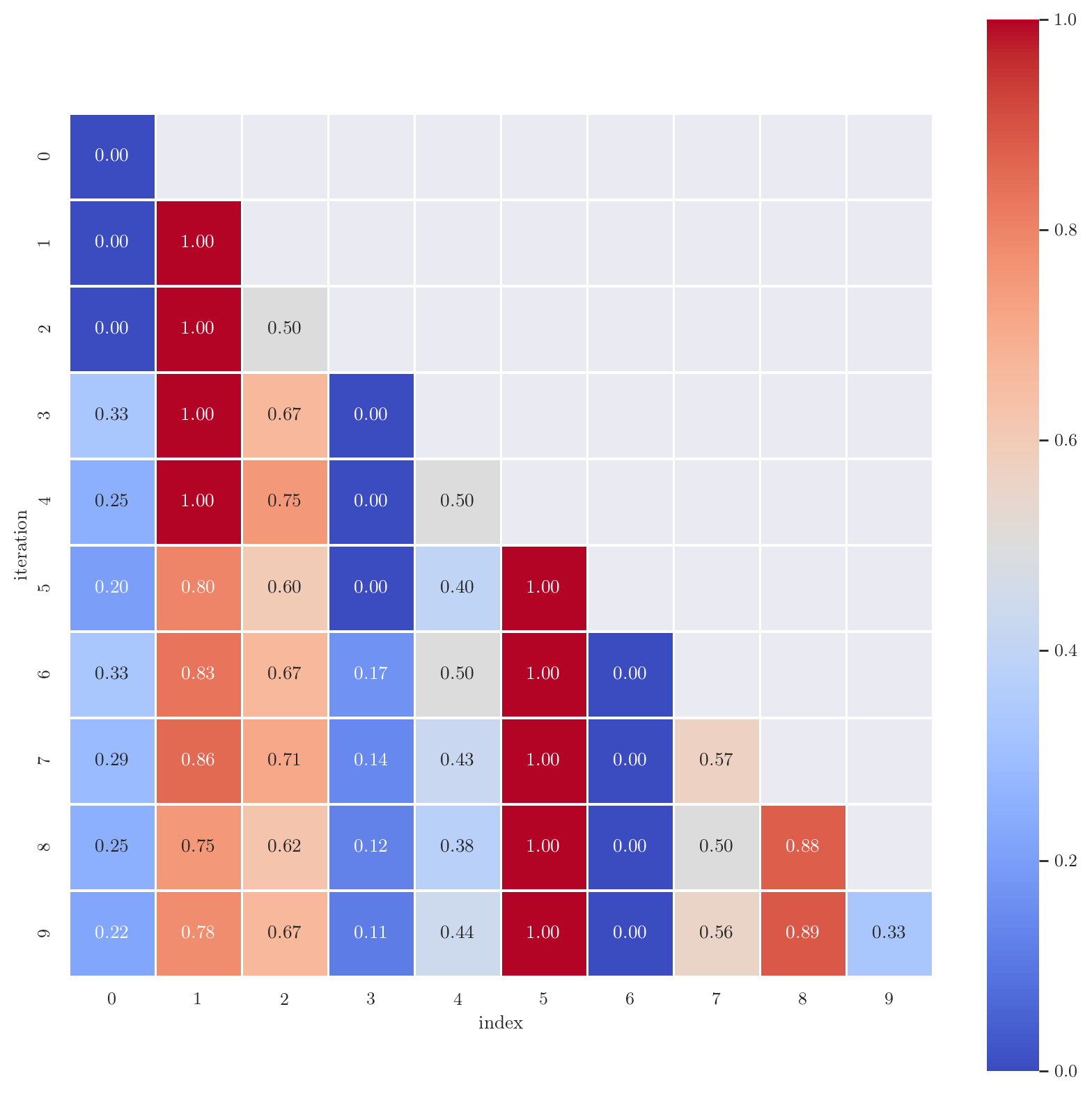}}}\\%
    \caption{Effect of prior $p(\boldsymbol{\theta})$ on class balance rate $\gamma$}%
    \label{fig:labels}%
\end{figure}

In this experiment we want to highlight the effect of prior $p(\boldsymbol{\theta})$ on class balance rate $\gamma$ in the BO loop resulting in re-shuffling of quantiles. Let's consider that the class-balance rate $\gamma$ does not change during the optimization process, therefore at each iteration, $\x_{\star}$ is added to the dataset $\mathcal{D}_n = \{\x_n, \y_n\}$, creating a shift in the ranking, hence a change in the quantile of observed $y$ values. \cite{tiao2021-bore} point out that this property can be exploited to make classifier training more efficient. Some strategies are well suited to reduce the classifier learning overhead and may include speeding up convergence such as \emph{early-stopping}, \emph{annealing} and \emph{importance sampling}. In fact, in the figure \ref{fig:labels}
 shows that after iteration $5$, referring to the empirical cumulative distribution of new candidates, we are \emph{almost} certain that the 
 dataset constructed include a \emph{local maximum} shown on the left panel, in the right panel we see clearly that the black-box function has attained its maximum, this property of the proposed method can help substantially reduce the classier learning overhead and speeding up convergence.

\section{Conclusion}
In this paper, we formulated a novel framework of Bayesian optimization problem that concerns uncertainty in density ratio estimation. Instead of trying to patch the deficiencies of the surrogate model, we reduce the Expected Improvement function to a probabilistic classification problem. We cast this observation in the Bayesian Neural Network setting, by seeking directly an approximate acquisition function, we therefore identified the link between the class-posterior probability and approximate density ratio estimation under \emph{Generalized Gauss Newton} method where we considered a linearized neural network. Further we demonstrate that our CPE-approach termed BOGGN outperforms against the state-of-the-art derivative-free global optimization methods under different variants of classifiers. The efficiency of our method leads to a fertile area of research in BO and approximate inference for future work.

\newpage
\bibliographystyle{apalike}
\bibliography{references}
\newpage
\section{Supplementary materials}
\subsection{Appendix.A : Approximate acquisition function}
\label{sec:Appendix.A}
We give the full derivation of approximate acquisition function, following \cite{bengio_bergstra_algoHPO}. In the case of the classification case, we use the expectation of the improvement utility function $U(\x,\y,\tau)$ over the posterior predictive distribution $p(\y|\z,\mathcal{D})$ where $\y_n \in \{0,1\}^C$. We have
\begin{align}
    \alpha(\x,\boldsymbol{\theta}, \mathcal{D}) &= \Ex_{p(\y|\x,\boldsymbol{\theta}, \mathcal{D})}\left[U(\x,\y,\tau)\right]\notag\\
    &=\int_{-\infty}^{\tau} p(\y|\x,\boldsymbol{\theta}, \mathcal{D})(\tau-\y)d\y\notag\\
    &=\int_{-\infty}^{\tau} \Ex_{p(\boldsymbol{\theta}| \mathcal{D})}\left[p(\y|\x,\boldsymbol{\theta}, \mathcal{D})\right](\tau-\y)d\y\label{eq:post_identy}\\
    &=\int_{-\infty}^{\tau}\int_{\boldsymbol{\theta}} p(\y|\x,\boldsymbol{\theta}, \mathcal{D})p(\boldsymbol{\theta}| \mathcal{D})(\tau-\y)d\y d\boldsymbol{\theta}\notag\\
    &=\frac{1}{p(\x|\boldsymbol{\theta}, \mathcal{D})}\int_{-\infty}^{\tau}\int_{\boldsymbol{\theta}} p(\y|\x,\boldsymbol{\theta}, \mathcal{D})p(\boldsymbol{\theta}| \mathcal{D})(\tau-\y)d\y d\boldsymbol{\theta}\label{eq:bayes_rule}\\
    &=\frac{1}{p(\x|\boldsymbol{\theta}, \mathcal{D})}\int_{-\infty}^{\tau}\int_{\boldsymbol{\theta}} p(\f(\x,\boldsymbol{\theta})|\y, \mathcal{D})p(\y|\mathcal{D})p(\boldsymbol{\theta}| \mathcal{D})(\tau-\y)d\y d\boldsymbol{\theta}\notag\\
    &=\frac{1}{p(\x|\boldsymbol{\theta}, \mathcal{D})}\int_{\boldsymbol{\theta}} \left[\tau \gamma \psi_1(\x,\boldsymbol{\theta})-\int _{-\infty}^{\tau}\y \psi_1(\x,\boldsymbol{\theta})p(\y|\mathcal{D})p(\boldsymbol{\theta}| \mathcal{D})d\y\right]d\boldsymbol{\theta}\label{eq:tau_cdf}\\
    &=\frac{1}{p(\x|\boldsymbol{\theta}, \mathcal{D})} \biggl(\Ex_{p(\boldsymbol{\theta}|\mathcal{D})}\left[\tau \gamma \psi_1(\x,\boldsymbol{\theta})\right]-\Ex_{p(\boldsymbol{\theta}| \mathcal{D})}\left[\int_{-\infty}^{\tau}\y \psi_1(\x,\boldsymbol{\theta})p(\y|\mathcal{D})d\y\right]\biggl)\notag\\
    &=\frac{1}{p(\x|\boldsymbol{\theta}, \mathcal{D})} \biggl\{\Ex_{p(\boldsymbol{\theta}|\mathcal{D})}\left[\psi_1(\x,\boldsymbol{\theta})\right]\left(\tau\gamma-\int_{-\infty}^{\tau}\y p(\y|\mathcal{D})d\y\right)\biggl\}\notag
\end{align}
where in equation (\ref{eq:bayes_rule}) we applied the Bayes'rule, in equation (\ref{eq:tau_cdf}) we have defined $\gamma=\Phi(\tau):=p(\y\leq\tau|\mathcal{D})$ and we used in equation (\ref{eq:post_identy}) the posterior identity
\begin{align*}
    p(\y|\x,\boldsymbol{\theta}, \mathcal{D})= \Ex_{p(\boldsymbol{\theta}| \mathcal{D})}\left[p(\y|\f(\x,\boldsymbol{\theta})\right].
\end{align*}
The denominator is evaluated to
\begin{align*}
    p(\x|\boldsymbol{\theta}, \mathcal{D})&= \int_{-\infty}^{\infty}p(\x|\boldsymbol{\theta},\y,\mathcal{D})p(\y|\x,\boldsymbol{\theta}, \mathcal{D})d\y\\
    &=\int_{-\infty}^{\infty}p(\x|\boldsymbol{\theta},\y,\mathcal{D})\Ex_{p(\boldsymbol{\theta}| \mathcal{D})}\left[p(\y|\f(\x,\boldsymbol{\theta})\right]d\y\\
    &=\int_{-\infty}^{\infty}\int_{\boldsymbol{\theta}}p(\x|\boldsymbol{\theta},\y,\mathcal{D})p(\y|\f(\x,\boldsymbol{\theta})p(\boldsymbol{\theta}| \mathcal{D})d\y d\boldsymbol{\theta}\\
    &=\Ex_{p(\boldsymbol{\theta}| \mathcal{D})}\left[\int_{-\infty}^{\tau}\psi_1(\x,\boldsymbol{\theta})p(\y|\boldsymbol{\theta},\mathcal{D}) d\y\right]+\Ex_{p(\boldsymbol{\theta}| \mathcal{D})}\left[\int_{\tau}^{\infty}\psi_2(\x,\boldsymbol{\theta})p(\y|\boldsymbol{\theta},\mathcal{D}) d\y\right]\\
    &=\Ex_{p(\boldsymbol{\theta}| \mathcal{D})}\left[\psi_1(\x,\boldsymbol{\theta})\gamma\right]+\Ex_{p(\boldsymbol{\theta}| \mathcal{D})}\left[\psi_2(\x,\boldsymbol{\theta})(1-\gamma)\right].
\end{align*}
Therefore, we can show that the approximate acquisition function is equivalent to the approximate $\gamma$-relative density ratio (\cite{yamada}) up to a constant factor,
\begin{align*}
    r_{\gamma}(\x,\boldsymbol{\theta},\mathcal{D},\tau) &= k\frac{\Ex_{p(\boldsymbol{\theta}| \mathcal{D})}\left[\psi_1(\x,\boldsymbol{\theta})\right]}{\Ex_{p(\boldsymbol{\theta}| \mathcal{D})}\left[\gamma\psi_1(\x,\boldsymbol{\theta})+\psi_2(\x,\boldsymbol{\theta})(1-\gamma)\right]}\\
    &\propto\Ex_{p(\boldsymbol{\theta}| \mathcal{D})}\left[\left(\gamma+(1-\gamma)\frac{\psi_2(\x,\boldsymbol{\theta})}{\psi_1(\x,\boldsymbol{\theta})}\right)^{-1}\right]\approx\hat{r}_{\gamma}(\x,\boldsymbol{\theta},\mathcal{D},\tau),
\end{align*}
where we recover an approximate formulation of the Expected Improvement (\cite{bengio_bergstra_algoHPO}) under $p(\boldsymbol{\theta}|\mathcal{D})$. Since $p(\boldsymbol{\theta})$ is intractable, we resort to approximation techniques, that is, we can approximate by a variational distribution $q(\boldsymbol{\theta})\approx p(\boldsymbol{\theta}|\mathcal{D})$.

\subsection{Appendix.B : Class-posterior probability}
\label{sec:Appendix.B}
We provide the class probability estimation according to the approximate $\gamma$-relative-density-ratio, this will establish the link between the class-posterior probability and the relative density-ration with respect to the approximate posterior distribution ${q(\boldsymbol{\theta})}$. First, we rewrite the approximate $\gamma$-relative density ratio and apply the Bayes' rule
\begin{align*}
     \hat{r}_{\gamma}(\x,\boldsymbol{\theta},\mathcal{D},\tau) 
    &\propto
    \Ex_{q(\boldsymbol{\theta})}\left[\frac{\psi_1(\x,\boldsymbol{\theta})}{\gamma\psi_1(\x,\boldsymbol{\theta})+\psi_2(\x,\boldsymbol{\theta})(1-\gamma)}\right]\\
    &=\Ex_{q(\boldsymbol{\theta})}\left[\frac{p(\x|\z=1,\y,\boldsymbol{\theta},\mathcal{D})}{\gamma p(\x|\z=1,\boldsymbol{\theta},\y,\mathcal{D})+(1-\gamma)p(\x|\z=0,\y,\boldsymbol{\theta},\mathcal{D})}\right]\\
    &= \Ex_{q(\boldsymbol{\theta})}\left[\frac{p(\z=1|\x,\boldsymbol{\theta},\mathcal{D})/p(\z=1)}{\gamma p(\z=1|\x,\boldsymbol{\theta},\mathcal{D}))/p(\z=1)+(1-\gamma)p(\z=0|\x,\boldsymbol{\theta},\mathcal{D})/p(\z=0)}\right]\\
   &= \Ex_{q(\boldsymbol{\theta})}\left[\frac{\pi(\x,\boldsymbol{\theta})/\gamma}{\gamma \pi(\x,\boldsymbol{\theta})/\gamma+(1-\gamma)(1-\pi(\x,\boldsymbol{\theta}))/(1-\gamma)}\right]\\
   &= \Ex_{q(\boldsymbol{\theta})}\left[\frac{\pi(\x,\boldsymbol{\theta})}{\gamma}\right],
\end{align*}
where $\pi(\x,\boldsymbol{\theta}):=p(\z=1|\f(\x;\boldsymbol{\theta}),\mathcal{D})$ denotes the likelihood of the model with respect to the hyperparameter $\boldsymbol{\theta}$.

\subsection{Appendix.C: The likelihood class-posterior}
\label{sec:Appendix.C}
In Bayesian learning setting, we wish to maximize the marginal likelihood, 
$p(\z|\f(\x;\boldsymbol{\theta})) :=\text{Bernoulli}(\z|\pi(\x;\boldsymbol{\theta}))$ denotes the likelihood of the model.
Next, we identify the dependency on marginal class-posterior probability, we expand this expression evaluated at \emph{maximum a priori} (MAP) $\boldsymbol{\theta}_{\star}$ to  
\begin{align}
     p(\z|\y,\f(\x;\boldsymbol{\theta}_{\star})) &= \frac{p(\y|\z, \f(\x; \boldsymbol{\theta}_{\star}))p(\z|\f(\x; \boldsymbol{\theta}_{\star}))}{p(\y|\X)}\notag\\
&=\frac{p(\y|\z=1, \psi_1(\x, \boldsymbol{\theta}_{\star}))p(\z=1|\f(\x, \boldsymbol{\theta}_{\star}))+p(\y|\z=0, \psi_2(\x, \boldsymbol{\theta}_{\star}))p(\z=0|\f(\x, \boldsymbol{\theta}_{\star}))}{p(\y|\X)}\notag\\
    &=\frac{p(\y|\z=1, \psi_1(\x, \boldsymbol{\theta}_{\star}))\pi(\x;\boldsymbol{\theta}_{\star})+p(\y|\z=0, \psi_2(\x, \boldsymbol{\theta}_{\star}))(1-\pi(\x,\boldsymbol{\theta}_{\star}))}{p(\y|\X)}\notag\\
    &=\frac{1}{p(\y|\X)}\left[\pi(\x,\boldsymbol{\theta}_{\star})\gamma+(1-\pi(\x,\boldsymbol{\theta}_{\star}))(1-\gamma))\right].
    \label{eq:objective_loss}
\end{align}
To compute the denominator, we derive the Laplace approximation to the marginal likelihood following \cite{mackay1992a}, that is, this method relies on a local quadratic approximation around the $\boldsymbol{\theta}_{\star}$, the log-joint distribution is given by
\begin{align*}
    \log p(\mathcal{D}, \boldsymbol{\theta}|\mathcal{M})&\approx 
    \log p(\y,\boldsymbol{\theta}_{\star}|\mathcal{M})-\left(\boldsymbol{\theta}-\boldsymbol{\theta}_{\star}\right)^T\mathcal{J}_{\boldsymbol{\theta}_{\star}}-\frac{1}{2}\left(\boldsymbol{\theta}-\boldsymbol{\theta}_{\star}\right)^T\mathcal{H}_{\boldsymbol{\theta}_{\star}}\left(\boldsymbol{\theta}-\boldsymbol{\theta}_{\star}\right)^T,
\end{align*}
where we denote  $\mathcal{H}_{\boldsymbol{\theta}}=-\nabla_{\theta \theta}^2 \log p(\mathcal{D}, \theta)$ and  $\mathcal{J}_{\boldsymbol{\theta}}=-\nabla_{\theta}\log p(\mathcal{D},\theta)$ as the Hessian and the gradient respectively of the \emph{map} objective. For a particular model $\mathcal{M}$, using the log-joint, we derive the Laplace approximation to the marginal likelihood:
\begin{align*}
    p(\mathcal{D}|\mathcal{M})&=\int p(\mathcal{D}, \boldsymbol{\theta}|\mathcal{M})d\boldsymbol{\theta}\\
    &\approx \int \text{exp}\left\{\log p(\mathcal{D}, \boldsymbol{\theta}_{\star})-(\boldsymbol{\theta}-\boldsymbol{\theta}_{\star})^T\mathcal{J}_{\boldsymbol{\theta}_{\star}}-\frac{1}{2}(\boldsymbol{\theta}-\boldsymbol{\theta}_{\star})^T\mathcal{H}_{\boldsymbol{\theta}_{\star}}(\boldsymbol{\theta}-\boldsymbol{\theta}_{\star})\right\}d\boldsymbol{\theta}:q(\mathcal{D}|\mathcal{}{M})\\
    &=p(\mathcal{D},\boldsymbol{\theta}_{\star})\int \text{exp}\left\{-\left(\boldsymbol{\theta}-\boldsymbol{\theta}_{\star}\right)^T\mathcal{J}_{\boldsymbol{\theta}_{\star}} -\frac{1}{2}(\boldsymbol{\theta}-\boldsymbol{\theta}_{\star})^T\mathcal{H}_{\boldsymbol{\theta}_{\star}}(\boldsymbol{\theta}-\boldsymbol{\theta}_{\star})\right\}d\boldsymbol{\theta}\\
    &=p(\mathcal{D},\boldsymbol{\theta}_{\star})\int \text{exp}\left\{-\frac{1}{2}\left(\boldsymbol{\theta}-\boldsymbol{\theta}_{\star}+\mathcal{H}^{-1}_{\boldsymbol{\theta}_{\star}}\mathcal{J}_{\boldsymbol{\theta}_{\star}}\right)^T\mathcal{H}_{\boldsymbol{\theta}_{\star}}\left(\boldsymbol{\theta}-\boldsymbol{\theta}_{\star}+\mathcal{H}^{-1}_{\boldsymbol{\theta}_{\star}}\mathcal{J}_{\boldsymbol{\theta}_{\star}}\right)+\frac{1}{2}\mathcal{J}_{\boldsymbol{\theta}_{\star}}\mathcal{H}^{-1}_{\boldsymbol{\theta}_{\star}}\mathcal{J}_{\boldsymbol{\theta}_{\star}}\right\}d\boldsymbol{\theta}\\
    &=p(\mathcal{D},\boldsymbol{\theta}_{\star})\int 
    \text{exp}\left\{\frac{1}{2}\mathcal{J}^{T}_{\boldsymbol{\theta}_{\star}}\mathcal{H}^{-1}_{\boldsymbol{\theta}_{\star}}\mathcal{J}_{\boldsymbol{\theta}_{\star}}\right\}
    \text{exp}\left\{-\frac{1}{2}\left(\boldsymbol{\theta}-\boldsymbol{\theta}_{\star}+\mathcal{H}^{-1}_{\boldsymbol{\theta}_{\star}}\mathcal{J}_{\boldsymbol{\theta}_{\star}}\right)^T\mathcal{H}_{\boldsymbol{\theta}_{\star}}\left(\boldsymbol{\theta}-\boldsymbol{\theta}_{\star}+\mathcal{H}^{-1}_{\boldsymbol{\theta}_{\star}}\mathcal{J}_{\boldsymbol{\theta}_{\star}}\right)\right\}d\boldsymbol{\theta}\\
    &=p(\mathcal{D},\boldsymbol{\theta}_{\star})\text{exp}\left\{\frac{1}{2}\mathcal{J}^{T}_{\boldsymbol{\theta}_{\star}}\mathcal{H}^{-1}_{\boldsymbol{\theta}_{\star}}\mathcal{J}_{\boldsymbol{\theta}_{\star}}\right\}
    (2\pi)^{P/2}\text{det}\left(\mathcal{H}_{\boldsymbol{\theta}_{\star}}\right)^{-1/2},
\end{align*}
and
\begin{align}
     p(\mathcal{D},\boldsymbol{\theta}_{\star}|\mathcal{M}) = \prod_{n}^N  p(\y_n|\psi_1(\x_n,\boldsymbol{\theta}_{\star})) p(\boldsymbol{\theta}_{\star})\label{eq:joint_likelihood},
 \end{align}
where $\psi_1(\x)$ is the linearized neural network at $\boldsymbol{\theta}_{\star}$. Finally to optimize the network parameters $\boldsymbol{\theta}$ we perform regular network training on the \emph{maximum a posteriori} (MAP) objective in equation (\ref{eq:objective_loss}) using stochastic optimizers like ADAM (\cite{kingma2017adam}) or SGD.
\subsection{Appendix.D: Optimizing the objective function}
\label{sec:Appendix.D}
Recall the objective function to be minimized takes the following form
\begin{align*}
   \mathcal{L}(\boldsymbol{\theta},\mathcal{D},\gamma|\mathcal{M}):=-\frac{1}{N}\biggl(\sum_{n=1}^N\log \pi(\x_n,\boldsymbol{\theta})\gamma+(1-\pi(\x_n,\boldsymbol{\theta}))(1-\gamma)
    -\sum_{n=1}^N\log q(\mathcal{D}_n|\mathcal{M})\biggr),
\end{align*}
To compute the approximate marginal distribution $q(\mathcal{D}_n|\mathcal{M})$, we make probabilistic prediction for new input $\x_{\star}$, in the case of the true posterior predictive $p(\boldsymbol{\theta}|\mathcal{D})$, we have
\begin{align*}
    p(\y_{\star}|\x_{\star},\boldsymbol{\theta},\mathcal{D}) = \Ex_{q(\boldsymbol{\theta}|\mathcal{D})}\left[p(\y_{\star}|\psi_1(\x_{\star},\boldsymbol{\theta}))\right],
\end{align*}
where we resort to posterior approximation techniques, in this paper we consider the Laplace approximation with \emph{Generalized-Gauss-Newton}, see section (\ref{sec:choice_q_approx}). Regardless of the posterior approximation chosen, we approximate this expectation by Monte Carlo sampling. In fact, the predictive can handle non-Gaussian likelihood and can depend on non-linearly $\f$ on $\boldsymbol{\theta}_s$. Therefore we estimate the approximate posterior $q(\boldsymbol{\theta})\approx p(\boldsymbol{\theta}|\mathcal{D})$ against the model likelihood $p(\mathcal{D|\boldsymbol{\theta}})$ such as
\begin{align}
    \Ex_{p(\boldsymbol{\theta}|\mathcal{D})}\left[p(\y|\psi_1(\x,\boldsymbol{\theta}))\right]\approx \frac{1}{S}\sum_{s=1}^Sp(\y|\psi_1(\x,\boldsymbol{\theta}_s))
    \label{eq:glm_predictive},
\end{align}
where $\boldsymbol{\theta}_s\sim q(\boldsymbol{\theta})=\Normal(\boldsymbol{\theta}_{\text{map}}, \Sigma_{\text{ggn}})$. Interestingly, since the Laplace-GGN posterior corresponds to the posterior of the linearized model, the Bayesian regression model turns to a \emph{generalized linear model} (GLM). 

\subsection{Equivalence of GLM formulation to Gaussian process}
The Bayesian GLM forumlation of the likelihood in equation (\ref{eq:joint_likelihood}) in weight space is equivalent to a Gaussian process (GP) in function space with a specific kernel (\cite{RasmussenW06}). We can write the GP-equivalent log-joint distribution as follow
\begin{align*}
    \sum_{n=1}^N \log p(\y_n|\f_n)+\log p(\f),
\end{align*}
where the GP prior is specified by its mean $\m(\x)$ and its covariance $\K(\x,\x^{\prime})$ under the prior $p(\boldsymbol{\theta})=\Normal(\m_0, \S_0)$ that takes the following form
\begin{align*}
    \m(\x) &= \Ex_{p(\boldsymbol{\theta})}[\psi_{\boldsymbol{\hat{\theta}}}(\x,\boldsymbol{\theta})]=\psi_{\boldsymbol{\hat{\theta}}}(\x,\m_0)\\
    \K(\x,\x^{\prime}) &= Cov_{p(\boldsymbol{\theta})}\left[\psi_{\boldsymbol{\hat{\theta}}}(\x,\boldsymbol{\theta}),\psi_{\boldsymbol{\hat{\theta}}}(\x^{\prime},\boldsymbol{\theta}))\right]\\
    &=\mathcal{J}_{\boldsymbol{\hat{\theta}}}(\x)\S_0\mathcal{J}_{\boldsymbol{\hat{\theta}}}(\x)^T,
\end{align*}
where $\boldsymbol{\hat{\theta}}=\boldsymbol{\theta}_{\text{\emph{map}}}$. In the case of a single output, the Laplace-GGN approximation to GP posterior $q(\f^{\star})$ at a new location $\x_{\star}$ is given by (\cite{RasmussenW06})
\begin{align*}
    \f_{\star}|\x_{\star},&\mathcal{D}\sim \Normal(\psi(\x_{\star},\boldsymbol{\theta}_{\text{\emph{map}}}),\boldsymbol{\Sigma}_{\star}^2)\\
    &\boldsymbol{\Sigma}_{\star}^2=\K(\x_{\star},\x_{\star})-\K(\x_{\star},\x_{N})\left(\K(\x_N,\x_N)+\boldsymbol{\Lambda}^{-1}(\y_N,\f_N)\right)^{-1}\K(\x_N,\x_{\star}),
\end{align*}
where $\K(\x_{\star},\x_N)$ denotes the kernel $\boldsymbol{\kappa}(.,.)$ evaluated between $\x_{\star}$ and the $N$ training points, and $\boldsymbol{\Lambda}(\y_N,\f_N)$ is a diagonal matrix. Analogoous to the GLM formulation in equation (\ref{eq:glm_predictive}), we obtain the GP predictive 
\begin{align*}
    p_{\mathcal{GP}}(\y|\x,\mathcal{D}) &= \Ex_{q(\f)}\left[p(\y|\f)\right]\\   
    &\approx \frac{1}{S}\sum_{s=1}^S p(\y|\f_s),
\end{align*}
where $\f_s\sim q(\f)$. This conversion from weight space to functional space allows further approximations, that is, since the parametric weight space posterior approximation sparsify the covariances of the parameters, the functional posterior approximation consider sparsity in data space. Therefore functional approximations of a GP model are orthogonal to parametric approximations in weight space.

\end{document}